\documentclass[3p,times,procedia,authoryear]{elsarticle}

\usepackage[colorinlistoftodos]{todonotes}

\usepackage{amsthm}
\usepackage{amsmath}
\usepackage{amssymb}
\usepackage{graphicx}
\usepackage{algorithm}
\usepackage{algorithmic,color}
\usepackage[unicode=true]{hyperref}
\usepackage{epstopdf}
\usepackage{multirow}
\usepackage{setspace}
\usepackage[subnum]{cases}
\usepackage{relsize}
\usepackage[colorinlistoftodos]{todonotes}

\usepackage{lineno}

\algsetup{indent=2em}

\begin{document}
\begin{frontmatter}

\title{A graph cut approach to  3D tree  delineation, using integrated airborne LiDAR and hyperspectral imagery }

\author{
  Juheon Lee$^{a,b}$, David Coomes$^{a*}$,     Carola-Bibiane Sch{\"o}nlieb$^{b}$ 
    Xiaohao Cai$^{a,b}$, Jan Lellmann$^{b}$, Michele Dalponte$^{a,c}$,
    Yadvinder Malhi$^{d}$, Nathalie Butt$^{d,e}$, Mike Morecroft$^{f}$, 
}

 \address{
  $^{a}$Forest Ecology and Conservation Group, Department of Plant Sciences,\\  
 University of Cambridge, CB2 3EA, UK \\
  $^{b}$Image Analysis Group, Department of Applied 
 Mathematics and Theoretical Physics (DAMTP),\\
 University of Cambridge, CB3 0WA, UK \\
 $^{c}$Department of Sustainable Agro-ecosystems and Bioresources, Research and Innovation Centre, Fondazione E. Mach, Via E. Mach 1, 38010 San Michele all'Adige (TN), Italy \\
 $^{d}$Environmental Change Institute, School of Geography and the Environment, \\ University of Oxford, OX1 3QY, UK \\ 
 $^{e}$Centre for Biodiversity and Conservation Science, The University of Queensland, \\ St Lucia, 4072, Qld, Australia 
 \\ $^{f}$Natural England, Cromwell House, 15 Andover Road, Winchester, SO23 7BT, UK
}
\cortext[cor1]{David Coomes,  dac18@cam.ac.uk}

\begin{abstract}
Recognising individual trees within remotely sensed imagery has important applications in  forest ecology and management. Several algorithms for tree delineation have been suggested, mostly based on locating local maxima or inverted basins in raster canopy height models (CHMs) derived from Light Detection And Ranging (LiDAR) data or  photographs. However, these algorithms often lead to inaccurate estimates of forest stand characteristics  due to the limited information content of raster CHMs. Here we develop a 3D tree delineation method which uses graph cut to delineate trees from the full 3D LiDAR point cloud, and also makes use of any optical imagery available (hyperspectral imagery in our case).  First,  conventional methods are used to locate local maxima in the CHM and generate an initial map of trees. Second, a graph is built from the LiDAR point cloud, fused with the hyperspectral data.  For computational efficiency, the feature space of hyperspectral imagery is reduced using robust PCA. Third, a  multi-class normalised cut is applied to the graph,  using the initial map of trees to constrain the number of clusters and their locations. Finally,  recursive normalised cut is used to subdivide, if necessary, each of the clusters identified by the initial analysis.  We call this approach Multiclass Cut followed by Recursive Cut (MCRC). The effectiveness of MCRC was tested using three datasets: i) NewFor, which includes several sites in the Alps and was established for comparing segmentation algorithms, ii) a coniferous forest in the Italian Alps, and iii) a deciduous woodland in the UK.  The performance of MCRC was usually superior to that of other delineation methods, and  was further improved by including  high-resolution optical imagery.  Since MCRC delineates the entire LiDAR point cloud in 3D,  it allows individual crown characteristics to be measured. MCRC is computationally demanding and, like current CHM-based approaches, is unable to detect understory trees. Nevertheless, by making full use of the data available, graph cut has the potential to considerably improve the accuracy of tree delineation.
\end{abstract}

 

\end{frontmatter}
 
 \section{Introduction}\label{sec:introduction}
 
There is much interest in using remote sensing to map individual tree crowns (ITCs) \citep{heinzel2012investigating,dalponte2014tree}  and measure various attributes of the identified trees. Until recently, applications of  remote sensing data to vegetation monitoring have focussed mostly on producing  rasterised  2D maps, with each pixel summarizing information from the many individual plants within them \citep{clark2005hyperspectral,asner2008airborne}. For example, NASA's Landsat 8 satellite produces imagery at 30m spatial resolution, which is too coarse to detect individual trees, but has been used to map forest types, deforestation,  blah blah among others [[ Hansen ]] and pixel-based approaches to analysing airborne remote sensing are already integrated into a national forest inventory program (e.g. Finland \citep{tomppo1993multi}).  However,  tree-centric approaches have the potential to advance forest research, by keeping track of individual  responses to pest and pathogen outbreaks, selective logging, fire, invasive species and climate change \citep{asner2008invasive,andersen2014monitoring,van2014predicting}.   Earth observation technology is producing information at increasingly high spatial resolution,  making ITC approaches an attractive alternative to pixel-based methods. In particular, airborne LiDAR produces a 3D point cloud indicating where laser pulses emitted from the transceiver have reflected off leaves, branches and the forest ground, making it possible to map individual trees over tens of thousands of hectares \citep{chen2006isolating,heinzel2012investigating,colgan2012mapping}.  In addition, airborne hyperspectral imagery can be used to estimate the physical and chemical properties of canopies, and when used alongside LiDAR can map these properties at an individual tree level \citep{asner2007carnegie}.  This "spectranomic"  approach has been used to map invasive tree species in Hawaiian rainforests \citep{asner2008pnas,asner2008invasive} and to quantify the spatial variation in biochemical diversity in tropical regions \citep{asner2008airborne,asner2011canopy}.  

Tree-centric approaches are  not widely adopted by the forest remote sensing community as  yet, in part because of problems with accuracy.  Classical delineation approaches work with rasterized canopy height models (CHMs) derived from the LiDAR point cloud. These methods include watershed algorithms \citep{chen2006isolating,koch2006detection,kwak2007detection,yu2011predicting},   variable window filtering \citep{hyyppa2001segmentation,solberg2006single},  multi-scale edge segmentation \citep{brandtberg2003detection}, and attentive vision methods \citep{palenichka2013multi}. All these approaches share several problems: (a) the smoothing process determines the number of trees detected in the CHM: too strong a smoothing factor leads to under-segmentation, while  too weak a factor generates many tree-like artifacts\citep{maltamo2014forestry}; (b)  sub-canopy trees are impossible to detect as they all rely solely on canopy surface geometry; and (c) the interpolation and smoothing processes involved in  generating CHMs result in underestimation of tree heights, meaning that additional post-processing is needed to rectify the results \citep{solberg2006single,koch2006detection}.  In order to address  these problems, more advanced methods that exploit the entire LiDAR point cloud have been developed. These methods include $k$-mean clustering \citep{morsdorf2004lidar}, normalised cut (NC) \citep{shi2000normalized,von2007tutorial,reitberger20093d,yao2012tree}, adaptive clustering \citep{lee2010adaptive}, support vector machine (SVM) \citep{secord2007tree,zhao2011characterizing}, and exploiting the spacing between top of trees \citep{li2012new}. Most of these approaches were  developed for managed coniferous forests, which are relatively straightforward to delineation because conical crowns have well defined peaks and forest size structure is simple. Benchmark datasets available to compare approaches also focus on coniferous forests \citep{newfor2012alpine}. Much less work has been done in tropical or temperate broadleaf forests, where intermingled dome-shaped crowns make delineation more challenging  \citep{reitberger20093d,yao2012tree,li2012new,newfor2012alpine,heinzel2012investigating,immitzer2012tree,colgan2012mapping}.  The  approach of Duncanson $et$ $al.$ (2014) hold promise in this regard;  they first delineated trees in the upper canopy using a watershed approach, then searched for "troughs" in the vertical structure of the local 3D point cloud that allowed them to strip away the taller trees and use the watershed algorithm a second time, to delineate subcanopy trees \citep{duncanson2014efficient}.    
In principle, the fusion of high resolution optical imagery with LiDAR data should lead to improvements in ITC delineation  \citep{chen2006isolating,koch2006detection,kwak2007detection,hyyppa2001segmentation,solberg2006single,brandtberg2003detection,dalponte2011system,yu2011predicting} by helping to distinguish neighbouring trees through differences in their radiometric properties \citep{kaasalainen2009radiometric,korpela2010range,korpela2010tree}. Aerial photographs and  multi / hyperspectral imagery could all be used for this purpose, as long as their spatial resolution is high enough (i.e. the pixel size is smaller than the minimum crown size that we need to detect) \citep{koch2010status,suarez2005use,holmgren2008species,breidenbach2010prediction,colgan2012mapping,heinzel2012investigating,dinuls2012tree,jakubowski2013delineating}. However, multi-sensor approaches are only possible if the different data are accurately co-aligned, thus image registration must be applied prior to their fusion (see \citep{dawn2010remote,le2011image}). A second issue is that extracting feature information directly from high dimensional data - such as the hyperspectral datasets - often leads to inaccurate results \citep{dalponte2008fusion}. Therefore, dimensional reduction is required before applying any delineation algorithm,  using feature extraction techniques such as  principal component analysis (PCA) \citep{candes2011robust}, or  by selecting influential features  from the original bands [[check]] \citep{dalponte2008fusion,dalponte2011system}.

This study seeks to overcome some  of the issues associated with  ITC delineation, by developing a new approach graph cut approach, based on Normalised Cut \citep{shi2000normalized,von2007tutorial,reitberger20093d,yao2012tree},  which combines optical and LiDAR data.  Normalised Cut is a well-established approach for grouping  points and/or pixels into disjoint clusters.  It starts with a matrix of similarity measures between all possible pairs of points and/or pixels, and uses the eigenvectors of that matrix to distinguish groups  \citep{shi2000normalized}.   In the case of LiDAR data, the similarity matrix is derived from the physical distance between points (nodes) in 3D space. In the case of hyperspectral data, the matrix is derived from their radiometric similarity and physical distances between pixels (nodes) in 2D space. NC seeks to partition the graph into clusters with high similarities between the nodes of the same clusters and a low similarities between nodes from different clusters. The advantage of the Normalised Cut approach is that graph weights can be defined using optical imagery alongside LiDAR, thus providing a framework for fusing different types of remote sensing datasets.  

Our main objective is to describe, and evaluate, a graph cut approach that can be used to delineate ITCs directly from a  3D LiDAR point cloud, using supplementary information for optical sensors.  To do this we (a) describe the data processing pipeline, including an efficient way of fusing LiDAR data and optical imagery and  various graph cut methods ; (b) examine the capability of MCRC to detect understory trees  and correctly segment canopy trees by  working with forest plot data from coniferous and broadleaf woodlands.  The paper is organized as follows: in Section \ref{sec:pre}, the general mathematical principles of the normalised cut approach are outlined.  The application of these principles to tree delineation in woodlands is  introduced in Section \ref{sec:method}. The test datasets used to exemplify  our approach are described in Section \ref{sec:data}. The performance of our approach is evaluated in Section \ref{sec:results}. Section \ref{sec:discussion} discusses MCRC  in relation to other approaches, and gives recommendations for future work. 

 \section{The general principles of the normalised cut approach } \label{sec:pre}

This section provides a formal outline of the normalised cut approach  \citep{shi2000normalized,reitberger20093d}.  A graph $\boldsymbol{G}$ is a pair of sets, $\boldsymbol{G} = (V,\boldsymbol{\epsilon})$,  where $V$ is the set of \textit{N} vertices and $\boldsymbol{\epsilon}$ is the set of edges.  Each edge $w_{ij} \in \boldsymbol{\epsilon}$ corresponds to a non-negative similarity weight between two vertices $i, j \in V$.  The objective of binary graph cut is to partition the graph into two disjoint sets $A$ and $B$ by cutting edges that connect the two sets, such that $ A\cup B =V \text{ and } A\cap B = \emptyset$.   We define the {\it cut} as  the sum over the weights of all edges that connect $A$ and $B$, that is
\begin{align} 
cut(A,B) = \sum_{i \in A, j \in B}w_{i,j} 
\end{align}
We define $assoc(A, V)$  to be the total weight of connections from nodes in $A$ to all nodes in the graph (i.e., $  \sum_{i \in A, j\in V} w_{ij}$). The normalised cut method finds sets $A$ and $B$ by minimising the following energy term: 
\begin{align} \label{eq-nc}
Ncut(A,B) = \frac{cut(A, B) }{assoc(A, V)} +  \frac{cut(A, B) }{assoc(B, V)}. 
\end{align}
In order to solve \eqref{eq-nc},  to find solution  $\boldsymbol{x} \in \mathbb{R}^N$ ,  we reformulate it as:
\begin{equation} \label{eq:standard}
\begin{aligned} 
\min_{\boldsymbol{x}\in \mathbb{R}^N} &
\boldsymbol{x}^T\boldsymbol{D}^{-\frac{1}{2}}(\boldsymbol{D}-\boldsymbol{W})\boldsymbol{D}^{-\frac{1}{2}}\boldsymbol{x} \\
\text{s.t.} & \quad
 \boldsymbol{x}^T\boldsymbol{D}^{\frac{1}{2}}\boldsymbol{1} = \boldsymbol{0},  \  \  \boldsymbol{x}^T\boldsymbol{x} = 1,
\end{aligned}
\end{equation}
where $\boldsymbol{D} \in \mathbb{R}^{N\times N}$ is a diagonal matrix with diagonal entries $d_{i}=\sum_{j=1}^N{w_{ij}}$,    $\boldsymbol{W} \in \mathbb{R}^{N\times N}$ is a symmetric matrix with entities $w_{ij}$, and  $\boldsymbol{1} \in \mathbb{R}^N$ is an all-ones vector.   Solutions are found by calculating the  eigenvectors of  matrix $\boldsymbol{D}^{-\frac{1}{2}}(\boldsymbol{D}-\boldsymbol{W})\boldsymbol{D}^{-\frac{1}{2}}$ .   The smallest eigenvector  is  $\boldsymbol{D}^{\frac{1}{2}}\boldsymbol{1}$ , which is a trivial solution,  and  is ignored.  It is the second smallest eigenvector that is taken as  the solution.   $V$ is  then split into two sets by thresholding the solution $\boldsymbol{x}$ ,  for example by taking the mean value of $\boldsymbol{x}$.

	The approach can be extended to search for multiple classes, by applying the binary graph cut  recursively (i.e.   sets A and B are further subdivided into  four sets,  and each of them may be  subdivided, and further subdivided) until the process is terminated by a stopping rule. The decision as to whether or not to make a split depend on whether the $Ncut$ energy value exceeds some predetermined threshold \citep{shi2000normalized}. By this recursive application of graph cut, the individual trees in the forest can be delineated.    However, the recursive scheme is computationally inefficient because it needs to solve equation \eqref{eq:standard} repeatedly until it reaches this predefined threshold. Since LiDAR data contains millions of points per hectare, recursive graph cut requires huge computational power to work on datasets larger than a few square metres. A further issue with the recursive approach is that equation \eqref{eq:standard} uses only the second smallest eigenvector \citep{shi2000normalized,von2007tutorial}, discarding information from subsequent eigenvectors that could help refine the partitioning. Finally, it is difficult to incorporate priors (i.e. initial guesses of the location of clusters) using this approach, which turns out to be important when delineating trees (see later).  
    For these reasons, there are advantages to using a multiclass  normalised cut  approach,  that searches for a predetermined number of classes, instead of using recursive binary cut  \citep{von2007tutorial}.  The multiclass problem can be understood as follows: let the solution matrix $\boldsymbol{X} = (\boldsymbol{x}_1, \cdots, \boldsymbol{x}_C) \in \mathbb{R}^{N \times C}$ where $C$ be the number of clusters. Then, the multiclass problem can be expressed in a similar way to problem \eqref{eq:standard}:
\begin{equation}\label{eq:mcstd}  
\begin{aligned} 
\min_{\boldsymbol{X}\in \mathbb{R}^{N\times C}} &
tr\big(\boldsymbol{X}^T\boldsymbol{D}^{-\frac{1}{2}}(\boldsymbol{D}-\boldsymbol{W})\boldsymbol{D}^{-\frac{1}{2}}\boldsymbol{X}\big) \\
\text{s.t.} & \quad
 \boldsymbol{x}_i^T\boldsymbol{D}^{\frac{1}{2}}\boldsymbol{1} = \boldsymbol{0},  \  \  \boldsymbol{x}_i^T\boldsymbol{x}_i = 1, \ \ i=1, \ldots, C,
\end{aligned}
\end{equation}
where  $V$ is split into $C$ sets by either $k$-means or spectral rotation. This  approach is computationally efficient, since the number of clusters is fixed at  $C$, equation \eqref{eq:mcstd} needs to be solved only once. 
	 However,  multiclass  normalised cut   has two problems when applied to forests.  The first problem is that  the number of trees has to be set in advance, which somewhat defeats the purpose of tree delineation!   This issue is resolved by taking each of the clusters identified by MC and applying a binary normalised cut to them recursively  to identify further trees within the cluster.   The second issue is that preliminary trials showed that MC performs poorly unless the algorithm is given some clues as to the whereabouts of trees.   To resolve this  problems, we first estimate the locations of tree tops from the local maxima of the CHM and use these locations as priors,  providing method \eqref{eq:mcstd} with an estimate of the number of clusters and their positions. Constrained normalised cut has been proposed by \citep{hu2013multi} but has never been used for ITCs delineation. This scheme regards a prior as an additional constraint to the solution of \eqref{eq:mcstd}, minimising  cut energy but also satisfying the condition that the correlation between the solution and the prior is larger than or equal to a predefined value ($\kappa$).  

Formally, let $\boldsymbol{S} = (\boldsymbol{s}_1, \cdots, \boldsymbol{s}_C) \in \mathbb{R}^{N \times C}$ be  a matrix of priors.  Then the MultiClass Normalised Cut with Priors (MC) approach is given by
\begin{equation}\label{eq:mcnc}
\begin{aligned} 
\min_{\boldsymbol{X}\in \mathbb{R}^{N\times C}} &
tr\big(\boldsymbol{X}^T\boldsymbol{D}^{-\frac{1}{2}}(\boldsymbol{D}-\boldsymbol{W})\boldsymbol{D}^{-\frac{1}{2}}\boldsymbol{X}\big) \\
\text{s.t.} & \quad
 \boldsymbol{x}_i^T\boldsymbol{D}^{\frac{1}{2}}\boldsymbol{1} = \boldsymbol{0},  \  \  \boldsymbol{x}_i^T\boldsymbol{x}_i = 1,   \  \ 
 \boldsymbol{x}_i^T\boldsymbol{s}_i \ge \kappa, \ \ i=1, \ldots, C,
\end{aligned}
\end{equation}
where $\kappa$ is a correlation parameter.  The solution of equation \eqref{eq:mcnc} gives $C$ separate clusters of data.  The correlation term is a hard constraint, which must be satisfied. In other words, the solution must have $C$ non-empty disjoint clusters. This method is much faster and more efficient than solving binary clustering recursively because  the number of clusters is fixed and  equation \eqref{eq:mcnc} is solved just once.    

\section{Methods} \label{sec:method}

\begin{figure}
\begin{center}
\begin{tabular}{c}
\includegraphics[width=80mm, height=80mm]{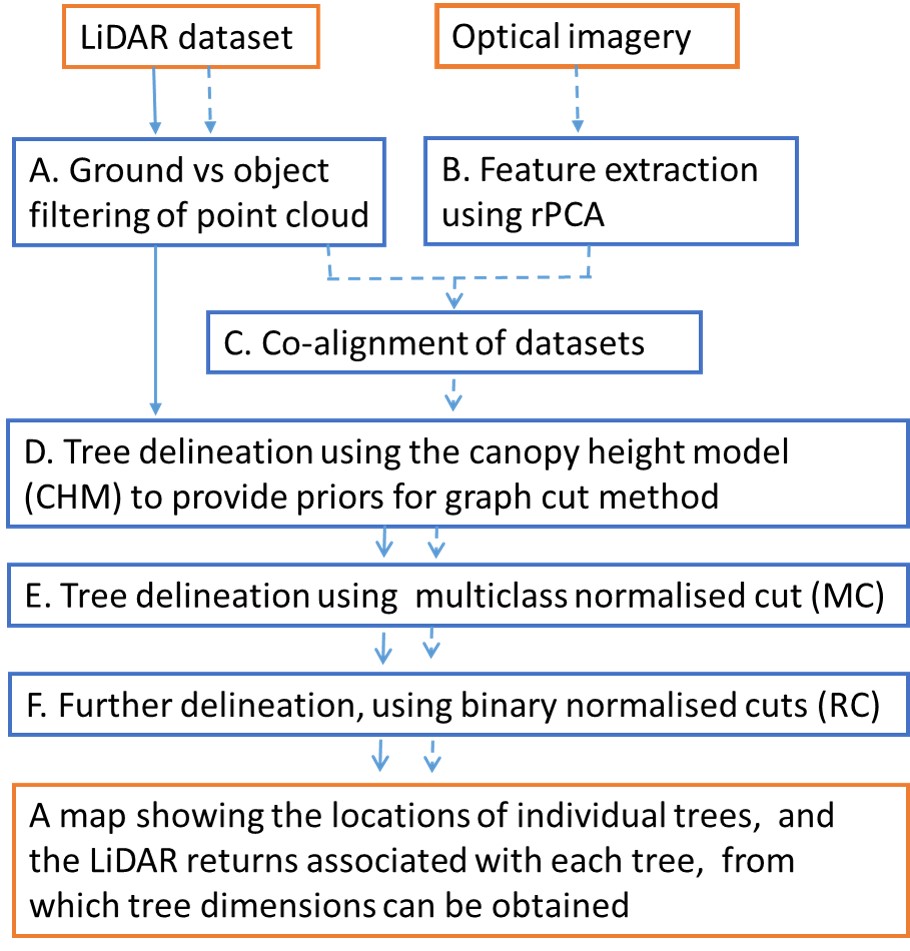}
\end{tabular}
\end{center}
\caption{The workflow used to delineate  individual tree crown from LiDAR data (solid line), and from LiDAR data fused with  optical imagery (dashed line).}
\label{fig:frame}
\end{figure}

The data processing pipeline shown in Figure \ref{fig:frame} has six steps: A. LiDAR data is separated into ground returns from which a digital elevation model (DEM) is constructed,  and  object returns,  from which a canopy height model (CHM) is  constructed;  B. if optical imagery is available, a state-of-the-art feature reduction method - robust PCA (rPCA) \citep{candes2011robust} - is used to reduce the number of hyperspectral features within the co-aligned dataset, to speed up processing;  C. if optical imagery is available,   LiDAR and optical imagery are registered (precisely co-aligned) using the NGF-Curv method that we  developed previously \citep{lee2015nonparametric};   D. a conventional  delineation approach based on the CHM is used to identify likely locations of upper-canopy trees;  E.  These locations are as priors a multiclass normalised cut (MC);  F.  Each of the clusters recognised by MC are subjected to recursive binary cutting. We call this MCRC (MultiClass Normalised Cut with Priors followed by Recursive Normalised Cut). Note that this method delineates ITCs directly from the 3D LiDAR point cloud, so  ITCs are not influenced by interpolation or smoothing errors prevalent in CHM-based approaches.  In the following section we explain each step in Figure \ref{fig:frame} in more detail.

\textbf{A. Ground vs object filtering of point cloud}
We performed initial modelling of terrain and canopy heights from the liDAR datasets using “Tiffs” 8.0: Toolbox for LidAR Data Filtering and Forest Studies, which employs a computationally efficient,
25 grid-based morphological filtering method described by Chen $et$ $al.$ (2007). Outputs included filtered ground and object points, as well as digital terrain models (DTM) and canopy height models (CHM).

\textbf{B. Feature extraction use rPCA}
Hyperspectral imagery is information rich - one of our datasets has information collected from 361 contiguous wavebands. Using such a highly dimensional data in a graph cut is computationally intensive,  and making it practically difficult to exploit the information \citep{dalponte2008fusion}. To alleviate this problem, the rPCA feature reduction technique is used in order to reduce the high dimensionality features space to a few meaningful features. Conventional PCA is sensitive to noise in data. In contrast, rPCA is designed to robustly recover a low rank matrix $L$ from a corrupted measurement matrix $M$ , to leave a sparse matrix of outliers $S$ \citep{candes2011robust}.  rPCA can be represented as the following minimization problem:
\begin{equation*}\label{eq:poo}
 \min_{L, S} 
  \left \{rank(L) + \lambda \Vert S \Vert_0 \right \} \quad 
 \text{s.t.} \ \  M = L + S,
\end{equation*}
where $\Vert \cdot \Vert_0$ is the $l_0$-norm which imposes a sparsity property on $S$, $rank(\cdot)$ is the dimensions of vector spaces spanned by the columns or rows of a matrix, and $\lambda$ is a regularisation parameter. Since this optimisation problem is intractable, in general, the rank and the $l_0$-norm are usually replaced by  the nuclear norm $\Vert \cdot \Vert_\ast$ (sum of singular values) and the $l_1$-norm (sum of the absolute values of the whole entries) respectively. This results in the following:
\begin{equation} \label{eq:rpca}
 \min_{L, S} 
  \left \{\|L\|_* + \lambda \| S \|_1 \right \} \quad
 \text{s.t.}  \ \  M = L + S.
\end{equation}

This objective function is convex, so it can be solved by various convex optimisation algorithms. In this paper, the alternating direction method of multipliers was used  \citep{yuan2009sparse}. We extracted the low rank parts $L$ which corresponds to the principal components in classic PCA. The first principal component was ignored because it contained illumination information rather of useful features of ITCs \citep{tochon2015use}. The second to fifth principal components were extracted and assigned to corresponding LiDAR points by using horizontal geospatial coordinates. If there is more than one LiDAR point in a pixel of hyperspectral imagery, then all points in the pixel were assigned the same rPCA coefficient. 

\textbf{C. Registration of remote sensing datasets}
LiDAR data and hyperspectral imagery are not usually precisely co-aligned when delivered by the data provider. Camera direction, topography and lens distortion all affect the quality of hyperspectral imagery, and LiDAR boresight is usually more accurate than that of the hyperspectral sensor, so inaccuracies remain even after geometric correction. To co-align these data, registration of LiDAR and optical imagery can be conducted using NGF-Curv algorithm, as proposed in \citep{lee2015nonparametric}. This non-parametric registration method uses normalised gradient field similarity measures with curvature regularisation.  Compared with the traditional parametric registration methods (e.g., \citep{le2011image,li2009robust}), the NGF-Curv method can handle nonlinear distortion and co-align multi-sensor imagery without any ground control points. The details of this method are described in \citep{lee2015nonparametric} and references therein.

\textbf{D. Local maxima detection}
Local maxima within the LiDAR point cloud provide the prior information on tree locations in this paper. Those local maxima can be easily extracted from the rasterized CHM, using a moving window approach \citep{hyyppa2001segmentation} or a watershed approach \citep{chen2006isolating}. We used a marker-based watershed approach for tree top detection implemented in TIFFS \citep{chen2006isolating}, comparing it efficacy with other approaches using the NewFor benchmark dataset (see below). All LiDAR points within 0.7m radius of each local maximum were identified as belong to the same cluster. We used these clusters as priors, thus enforcing the solution of equation \eqref{eq:mcnc}. The marker-based watershed approach is just one of the possible methods to set up priors \citep{reitberger20093d} (see Section \ref{sec:results}).

\textbf{E. MultiClass Normalised Cut with Priors (MC)}
To build the graph for MC,  weights need to be assigned to the vertices, which are given by the LiDAR points. We used a normalised weight that is a function of Euclidean distances between vertices $i$ and $j$  in horizontal ($x,y$) and vertical ($z$) space, as well as the similarity of their  hyperspectral features ($fts$) :  
\begin{equation} \label{eq:RBF}  \mathlarger{
 w_{ij} = e^{\frac{\| (xy)_i - (x,y)_j \|}{\sigma_{xy}^2}} \times e^{\frac{\| z_i - z_j \|}{\sigma_{z}^2}}  
\times e^{\frac{\| fts_i - fts_j \|}{\sigma_{fts}^2}}},
\end{equation} 

where  bandwidth parameters $\sigma_{xy},\sigma_{z},\sigma_{fts}$  act to normalise the function, and are parameters selected by the user.  For constructing the graph, we observe that a fully connected graph requires $\mathcal{O}(n^2)$ memory complexity,  which is not practical. Instead, a $d$-neighbourhood sampling strategy is adopted, where weights are  computed only within a radius $d$ of a vertex.  In our examples, $d$ ranged from 0.5m to 2m depending on the point density of LiDAR (lower radii at higher densities to reduce the memory costs). Equation \eqref{eq:mcnc} was solved with a $d$-neighbourhood similarity matrix and pre-defined clusters taken from the local maxima.  The MC approach  segments the 3D LiDAR point cloud into the same number of tree crowns as identified by traditional CHM-based methods, because this information is used as a "hard" prior. It also  suffers from the same problems as classic approaches in terms of failing to detect understory trees. 

\textbf{F. Recursive Normalised Cut (RC)}
The  RC method \eqref{eq:standard} described in Section \ref{sec:pre} is effective at ITC delineation, including the detection of understory trees\citep{reitberger20093d}, but is computationally costly if applied to the whole dataset. For this reason,  we applied RC to each of the clusters obtained from an  initial MC, to provide an opportunity for canopy "trees" to be further subdivided and for subcanopy trees to be detected.  

\section{Dataset Description and Design of Experiments} \label{sec:data}

The accuracy of the MCRC algorithm was tested on (a) a set forest plots located in the Alps which form part of the NewFor benchmarking project, established specifically for the purpose of comparing ITC algorithms \citep{newfor2012alpine}, (b) a coniferous forest located near Trento in the Italian Alps, and (c) a lowland deciduous forest located near Oxford, UK.

\textbf{(a) The NewFor LiDAR Single Tree Detection Benchmark Dataset }consists of LiDAR and ground-truth information from 14 survey sites in the Alps (10 pilot areas in 6 countries) \citep{eysn2015benchmark}. A major advantage of working with the NewFor benchmark dataset is that it provides an objective means of comparing our approach with others, and includes sophisticated validation software with which to evaluate algorithms by matching ITCs derived from LiDAR with known tree locations in the field. The ground truth data were provided with geocoordinates, tree height, DBH and canopy volume information. The errors of geocoordinates were less than 1m. The LiDAR point density was more than 10 per $\text{m}^2$ in 12 out of 14 study site. The ranges of the LiDAR point density in the NewFor benchmark dataset were from 4 to 121 per $\text{m}^2$. A disadvantage of the NewFor dataset with regard to our proposed delineation procedure is that it does not contain any optical imagery (i.e. we worked with the pipeline shown with solid lines in Figure \ref{fig:frame}). Note that these datasets are primarily coniferous, which are relatively straightforward to delineate because conifers have distinct peaks to their crowns. 

\textbf{(b) The Italian Alps dataset} was collected from a location near Trento. It consists of hyperspectral imagery, LiDAR data and ground-based tree maps for 7 plots dominated by coniferous trees. Each plot is a circle of 15m radius. In these plots, all trees with DBH above 1cm were accurately georeferenced by differential GPS and manually corrected with local reference trees from LiDAR data. The estimated error of the ground truth of tree positions was 1m. The hyperspectral imagery were collected with an AISA Eagle sensor, covering 400--970nm with 61 spectral bands, while the LiDAR data were acquired by a Riegl LMS-Q680i sensor at an unusually high point density ($\geq$ 87 points per $\text{m}^2$). Hyperspectral data were collected on $13^\text{th}$ of June 2013, while LiDAR data were collected between $7^\text{th}$ and $9^\text{th}$ of September 2012.
 
\textbf{(c) The English broadleaf woodland dataset} was collected from Wytham Woods, Oxfordshire, England. It contains  hyperspectral and LiDAR data over 18 hectares of temperate woodland dominated by deciduous angiosperm species. The plot is fully mapped and all trees with DBH above 5cm are permanently tagged. As tree height was measured physically only for some selected samples we used allometric equations to estimate tree height for all the other trees. The estimated positioning error of the plot corners is approximately 2m, while tree positions are located within about 5m. Hyperspectral imagery  was collected in June, 2014 using an AISA Fenix sensor by the airborne research and survey facility of the national environmental research council of UK (NERC-ARSF). It covers 400--2500nm with 361 spectral bands. LiDAR data were collected by a Leica ALS-50 II scanner simultaneously with a AISA Fenix hyperspectral imagery. The LiDAR point density was 6 points per $\text{m}^2$. 

The optimal parameters for the MCRC were found by trial and error ( Table\ref{table:parameter}). 

\begin{table}[!h]
\centering 
\caption{Values of bandwidth parameters selected for the normalised cuts in three experiments}
\vspace{0.1in}
\begin{tabular}{ccccccc}
\hline
\multicolumn{1}{|c||}{\multirow{2}{*}{Dataset}}  & \multicolumn{3}{c|}{MC} & \multicolumn{3}{c|}{RC}\\ 
\cline{2-7}
\multicolumn{1}{|c||}{} & \multicolumn{1}{c|}{$\mathlarger \sigma_{xy}$} & \multicolumn{1}{c|}{$\mathlarger \sigma_{z}$} & \multicolumn{1}{c|}{$\mathlarger \sigma_{fts}$} & \multicolumn{1}{c|}{$\mathlarger \sigma_{xy}$} & \multicolumn{1}{c|}{$\mathlarger \sigma_{z}$} & \multicolumn{1}{c|}{$\mathlarger \sigma_{fts}$}\\ \hline\hline
 \multicolumn{1}{|l||}{Italian Alps} & \multicolumn{1}{c|}{1}& \multicolumn{1}{c|}{3} & \multicolumn{1}{c|}{0.005}& \multicolumn{1}{c|}{0.5} & \multicolumn{1}{c|}{2} & \multicolumn{1}{c|}{0.005} \\   \hline
 \multicolumn{1}{|l||}{NewFor benchmark} & \multicolumn{1}{c|}{2}& \multicolumn{1}{c|}{5} & \multicolumn{1}{c|}{n/a} & \multicolumn{1}{c|}{2} & \multicolumn{1}{c|}{5} & \multicolumn{1}{c|}{n/a} \\  \hline
 \multicolumn{1}{|l||}{English broadleafs}  & \multicolumn{1}{c|}{2}& \multicolumn{1}{c|}{3} & \multicolumn{1}{c|}{0.005} & \multicolumn{1}{c|}{2} & \multicolumn{1}{c|}{3} & \multicolumn{1}{c|}{0.005} \\  \hline
\end{tabular}
\label{table:parameter}
\end{table}

The validation of the ITC delineation was conducted using the tree matching software provided by the NewFor project \citep{eysn2015benchmark,kaartinen2012international}, which compares relative positions and heights of segmented trees with those recorded in the ground plots. Specifically, it measures 2D Euclidean distance and height difference between ground truth and segmented trees. Ground-truth trees within 5m of segmented trees, both  horizontally and vertically, were considered as  potential matches. The closest tree in both horizontal and vertical distances was selected as the match. By comparing not only tree positions but also heights, this validation software reduces errors arising from the inaccurate georeferencing of the ground truth. The sensitivity of the MCRC algorithm with respect to prior information was examined by comparing its results when the TIFFS watershed algorithm \citep{chen2006isolating} and moving window filtering (MWF) \citep{hyyppa2001segmentation} were used to establish the priors. In order to evaluate the performance of MCRC, we compared our segmentation approach with RC \citep{reitberger20093d} and the CHM-based watershed algorithm of TIFFS \citep{chen2006isolating}.

\section{Results}\label{sec:results}
\subsection{Tree delineation using LiDAR imagery}

\begin{figure*}[!htb]  
\begin{center} 
\begin{tabular}{cc}
\includegraphics[width=78mm, height=72mm]{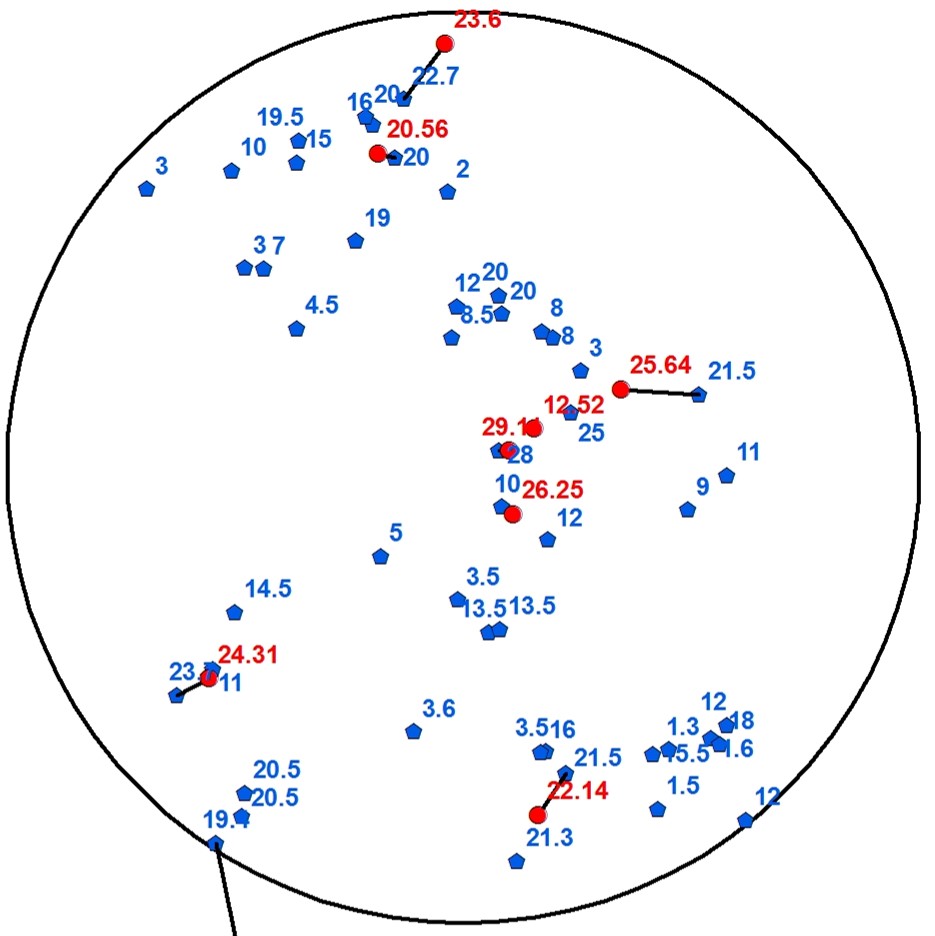}  &
\includegraphics[width=78mm, height=72mm]{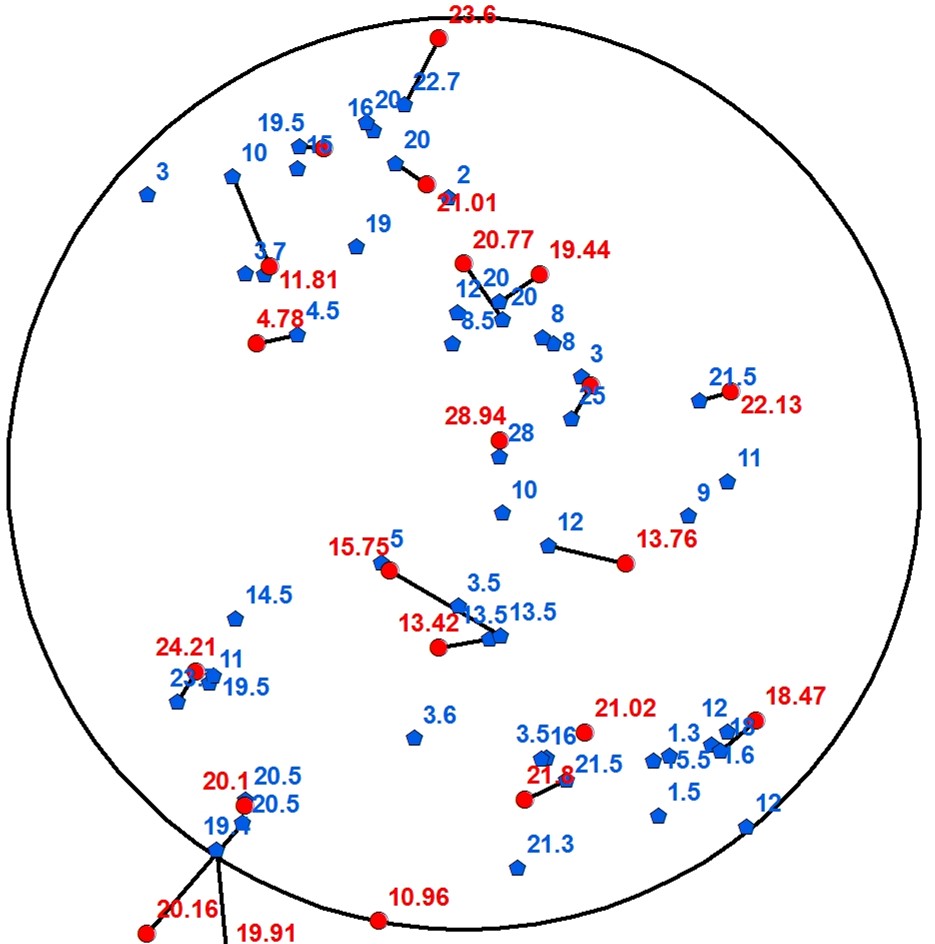} \\
 RC &  MCRC
\end{tabular}
\end{center}
\caption{Examples of MCRC tree delineation for a forest plot in the Italian Alps dataset, for which tree locations and sizes have been mapped on the ground (plot 77). Results are  projected onto a 2D plane: each delineated tree is shown as a red circle, and each tree measured on the ground is shown as a blue pentagon. The outer circle is the 15m-radius plot boundary. The numbers in red and blue colours indicate tree heights of segmented trees and ground truth, respectively. The dark solid line shows matches between segmented and ground-measured trees,  based on proximity and  height similarity.  It can be seen that many small trees are not detected}
\label{fig:2D}
\end{figure*}

\begin{figure*}[!htb]
\begin{center}
\begin{tabular}{cc}
\includegraphics[width=70mm, height=70mm]{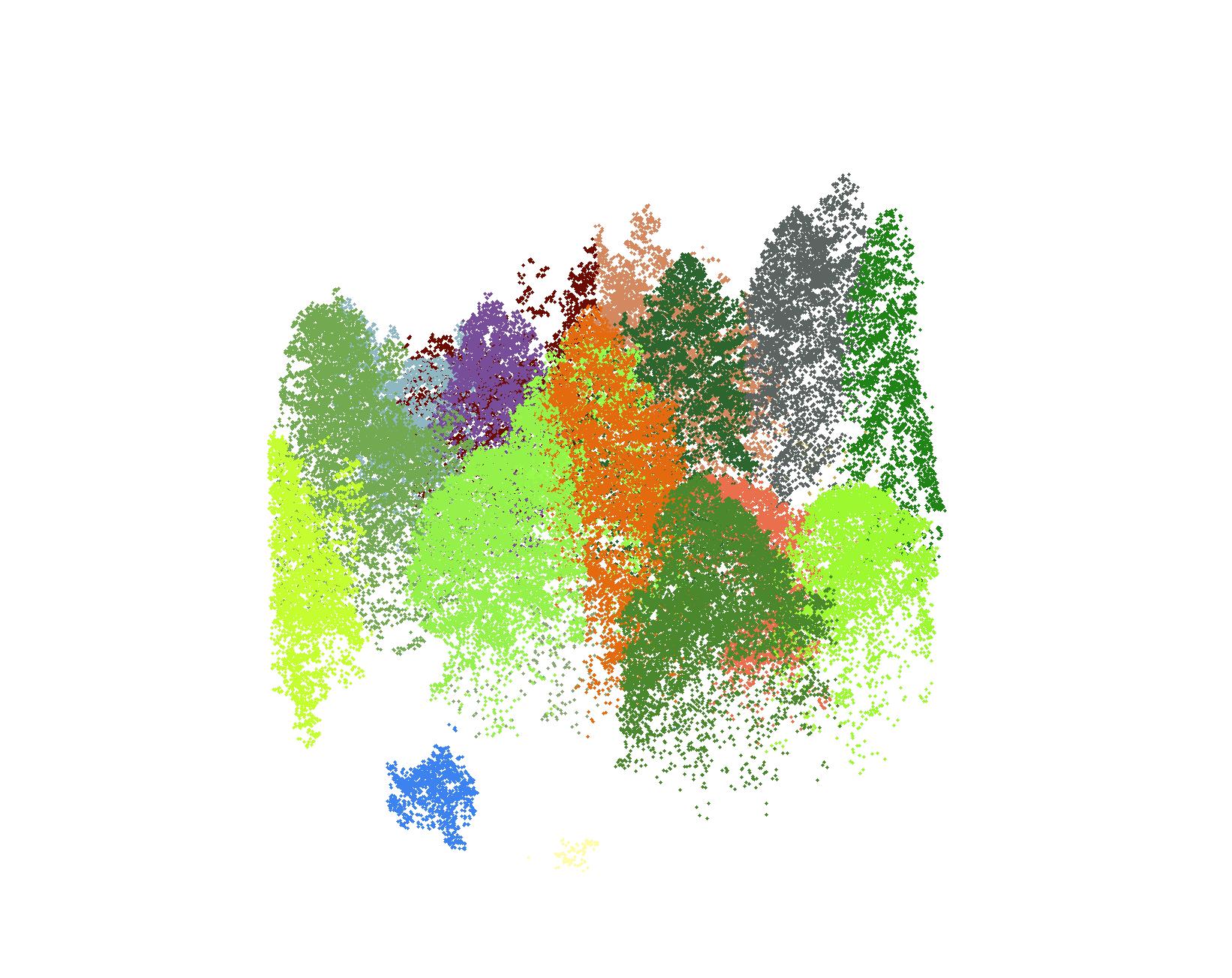}  &
\includegraphics[width=70mm, height=70mm]{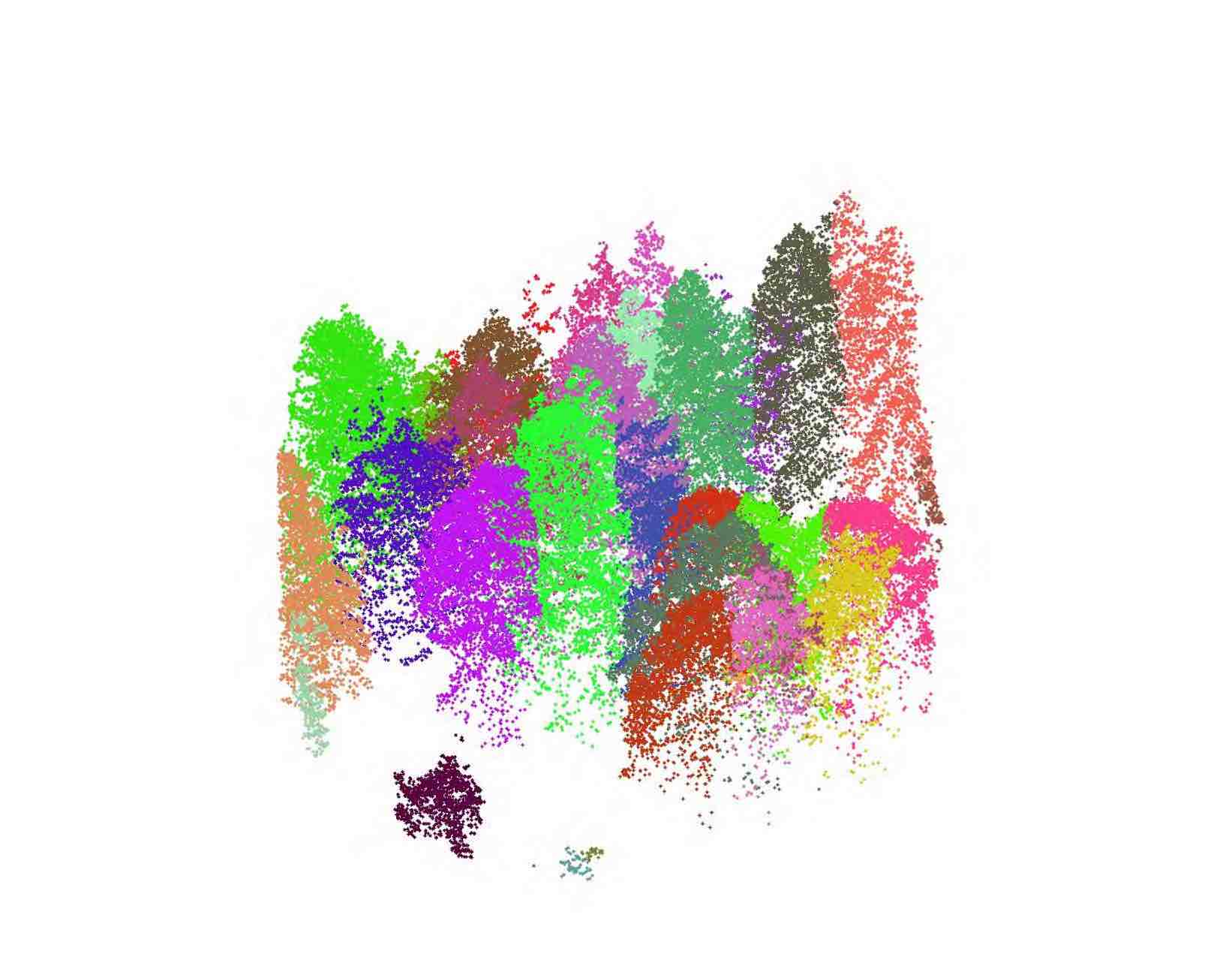} 
  \\
 RC &  MCRC\\
\end{tabular}
\end{center}
\caption{3D examples of individual tree delineation by RC (left) and MCRC (right) algorithms in the Italian dataset (plot 77).}
\label{fig:3D}
\end{figure*}

The performance of our graph cut algorithms was compared with that of the TIFFS watershed algorithm, which uses the canopy height model to find trees. We found that the performance of TIFFS was equal to, or surpassed, that of eight other methods already evaluated using the NewFor benchmark datasets \citep{newfor2012alpine,eysn2015benchmark}, and on that basis it was chosen as our point of comparison,  The TIFFS algorithm was also used to provide priors for the MCRC (i.e. step D in the pipeline shown in Fig 1). TIFFs was selected because the MCMC results were more accurate with TIFFs priors than with moving-window-filtering priors:  use of TIFFs led to slightly better performance in five out of seven Italian Alps test plots (i.e. plots 77, 102, 129, 220 and 292) and similar performance in the two remaining plots (Tables \ref{table:performance} and \ref{table:isub}).  
The MCRC approach performed better than RC alone, but many small trees were missed. Figures \ref{fig:2D} and \ref{fig:3D} illustrate the results of individual tree detection by MCRC versus RNC (in 2D and 3D respectively). RC detected correctly only 14\% (7 out of 50 trees) tree crowns in plot 77 of the Italian Alps dataset, while MCRC detected 34\%. Moreover, it can be seen in Figure \ref{fig:3D} that RC leads to unrealistic tree delineation. The performance of RC was poor in all the experiments we performed (results not shown), therefore we did not consider it further. The performances of TIFFS and MCRC in the Italian dataset were shown in Tables \ref{table:performance} and \ref{table:isub}, where MCRC showed slightly better performance to find understory trees compared to TIFFS. More precisely, MCRC algorithm outperformed TIFFS in five sites and the performance was the same in the other two test sites. The performance of delineation algorithms varied with height class bands (Table \ref{table:isub}). MCRC found a few more understory trees than TIFFS,  but its performance was still poor.  
 
\begin{table}[!h]
\centering 
\caption{Comparison of the performance of delineation algorithms  applied to seven forest plots in the Italian Alps.   MCRC is the normalised cut approach developed in this paper (see Figure 1), which can use LiDAR and hyperspectral data (Hyp) or just LiDAR data.  MCRC uses priors obtained by conventional approaches, which are given in brackets: a  watershed algorithm (TIFFS) and Moving Window Filter (MWF) were used to locate local maxima which are used as priors.  MCRC is compared with TIFFS.    `Ground Truth' is the number of stems ($>$ 1 cm DBH) recorded in the field plots.  `Extracted' means the number of trees delineated by the algorithms. while `matched' indicates the number of correctly segmented trees, assessed by the NewFor matching algorithm.}
\vspace{0.1in}
\footnotesize
\begin{tabular}{cccccccccc}
\hline
\multicolumn{1}{c||}{\multirow{2}{*}{Plot}} & \multicolumn{1}{c|}{Ground} & \multicolumn{2}{c|}{MCRC (MWF) } & \multicolumn{2}{c|}{TIFFS} & \multicolumn{2}{c|}{MCRC (TIFFS)} & \multicolumn{2}{c}{MCRC (TIFFS) Hyp}\\ 

\cline{3-10}
\multicolumn{1}{c||}{} &\multicolumn{1}{c|}{Truth} & \multicolumn{1}{c|}{Extracted} & \multicolumn{1}{c|}{Matched} & \multicolumn{1}{c|}{Extracted} & \multicolumn{1}{c|}{Matched} & \multicolumn{1}{c|}{Extracted} & \multicolumn{1}{c|}{Matched} & \multicolumn{1}{c|}{Extracted} & \multicolumn{1}{c}{Matched}
\\ \hline\hline

 \multicolumn{1}{c||}{Plot 77} &  \multicolumn{1}{c|}{50} & \multicolumn{1}{c|}{15}& \multicolumn{1}{c|}{15} & \multicolumn{1}{c|}{17} & \multicolumn{1}{c|}{15} & \multicolumn{1}{c|}{19} & \multicolumn{1}{c|}{17} & \multicolumn{1}{c|}{18} & \multicolumn{1}{c}{16}\\  
 \hline
 \multicolumn{1}{c||}{Plot 91} &  \multicolumn{1}{c|}{72} & \multicolumn{1}{c|}{23}& \multicolumn{1}{c|}{22} & \multicolumn{1}{c|}{25} & \multicolumn{1}{c|}{22} & \multicolumn{1}{c|}{25} & \multicolumn{1}{c|}{22} & \multicolumn{1}{c|}{25} & \multicolumn{1}{c}{22}\\ 
 \hline
 \multicolumn{1}{c||}{Plot 102} &  \multicolumn{1}{c|}{35} & \multicolumn{1}{c|}{18}& \multicolumn{1}{c|}{17} & \multicolumn{1}{c|}{17} & \multicolumn{1}{c|}{16} & \multicolumn{1}{c|}{20} & \multicolumn{1}{c|}{18} & \multicolumn{1}{c|}{19} & \multicolumn{1}{c}{18}\\ 
 \hline
 \multicolumn{1}{c||}{Plot 129} &  \multicolumn{1}{c|}{11} & \multicolumn{1}{c|}{10}& \multicolumn{1}{c|}{9} & \multicolumn{1}{c|}{14} & \multicolumn{1}{c|}{8} & \multicolumn{1}{c|}{14} & \multicolumn{1}{c|}{9} & \multicolumn{1}{c|}{15} & \multicolumn{1}{c}{9}\\ 
 \hline
 \multicolumn{1}{c||}{Plot 220} &  \multicolumn{1}{c|}{21} & \multicolumn{1}{c|}{17}& \multicolumn{1}{c|}{17} & \multicolumn{1}{c|}{18} & \multicolumn{1}{c|}{17} & \multicolumn{1}{c|}{19} & \multicolumn{1}{c|}{18} & \multicolumn{1}{c|}{19} & \multicolumn{1}{c}{18}\\ 
 \hline
 \multicolumn{1}{c||}{Plot 274} &  \multicolumn{1}{c|}{57} &\multicolumn{1}{c|}{31}& \multicolumn{1}{c|}{31} & \multicolumn{1}{c|}{32} & \multicolumn{1}{c|}{32} & \multicolumn{1}{c|}{32} & \multicolumn{1}{c|}{32} & \multicolumn{1}{c|}{32} & \multicolumn{1}{c}{32}\\ 
 \hline
 \multicolumn{1}{c||}{Plot 292} &  \multicolumn{1}{c|}{39} & \multicolumn{1}{c|}{15}& \multicolumn{1}{c|}{6} & \multicolumn{1}{c|}{12} & \multicolumn{1}{c|}{10} & \multicolumn{1}{c|}{12} & \multicolumn{1}{c|}{11} & \multicolumn{1}{c|}{14} & \multicolumn{1}{c}{11}\\  
 \hline \hline
 \multicolumn{1}{c||}{Overall} &  \multicolumn{1}{c|}{285} & \multicolumn{1}{c|}{129}& \multicolumn{1}{c|}{117} & \multicolumn{1}{c|}{135} & \multicolumn{1}{c|}{120} & \multicolumn{1}{c|}{141} & \multicolumn{1}{c|}{127} & \multicolumn{1}{c|}{142} & \multicolumn{1}{c}{126}\\    \cline{1-10}                    
\end{tabular}
\label{table:performance}
\end{table}

\begin{table}[!h]
\centering 
\caption{Comparison of the performance of delineation algorithms for different height bands of trees within the Italian Alps dataset . `Extract' means the number of trees delineated by the algorithms. `Match' is the number of trees that were matched to trees in mapped forest plots which had similar (x,y) coordinates and were of similar heights.  See Table 2 for explanation of model names}
\vspace{0.1in}
\footnotesize
\begin{tabular}{cccccccccc}
\hline
\multicolumn{1}{c||}{\multirow{2}{*}{Plot}} & \multicolumn{1}{c|}{Ground} & \multicolumn{2}{c|}{MCRC (MWF) } & \multicolumn{2}{c|}{TIFFS} & \multicolumn{2}{c|}{MCRC (TIFFS)} & \multicolumn{2}{c}{MCRC (TIFFS) Hyp}\\ 

\cline{3-10}
\multicolumn{1}{c||}{} &\multicolumn{1}{c|}{Truth} & \multicolumn{1}{c|}{Extracted} & \multicolumn{1}{c|}{Matched} & \multicolumn{1}{c|}{Extracted} & \multicolumn{1}{c|}{Matched} & \multicolumn{1}{c|}{Extracted} & \multicolumn{1}{c|}{Matched} & \multicolumn{1}{c|}{Extracted} & \multicolumn{1}{c}{Matched}
\\ \hline\hline

 \multicolumn{1}{c||}{$h \geq 20\text{m}$} &  \multicolumn{1}{c|}{137} & \multicolumn{1}{c|}{114}& \multicolumn{1}{c|}{103} & \multicolumn{1}{c|}{120} & \multicolumn{1}{c|}{108} & \multicolumn{1}{c|}{125} & \multicolumn{1}{c|}{111} & \multicolumn{1}{c|}{125} & \multicolumn{1}{c}{111}\\  
 \hline
 \multicolumn{1}{c||}{$ 15\text{m}\leq h < 20\text{m}$} &  \multicolumn{1}{c|}{29} & \multicolumn{1}{c|}{7}& \multicolumn{1}{c|}{6} & \multicolumn{1}{c|}{8} & \multicolumn{1}{c|}{7} & \multicolumn{1}{c|}{8} & \multicolumn{1}{c|}{8} & \multicolumn{1}{c|}{8} & \multicolumn{1}{c}{7}\\ 
 \hline
 \multicolumn{1}{c||}{$ 10\text{m}\leq h < 15\text{m}$} &  \multicolumn{1}{c|}{33} & \multicolumn{1}{c|}{5}& \multicolumn{1}{c|}{5} & \multicolumn{1}{c|}{4} & \multicolumn{1}{c|}{3} & \multicolumn{1}{c|}{6} & \multicolumn{1}{c|}{6} & \multicolumn{1}{c|}{7} & \multicolumn{1}{c}{6}\\ 
 \hline
 \multicolumn{1}{c||}{$ 5\text{m}\leq h < 10\text{m}$} &  \multicolumn{1}{c|}{36} & \multicolumn{1}{c|}{3}& \multicolumn{1}{c|}{3} & \multicolumn{1}{c|}{2} & \multicolumn{1}{c|}{2} & \multicolumn{1}{c|}{2} & \multicolumn{1}{c|}{2} & \multicolumn{1}{c|}{2} & \multicolumn{1}{c}{2}\\ 
 \hline
 \multicolumn{1}{c||}{$ 2\text{m}\leq h < 5\text{m}$} &  \multicolumn{1}{c|}{24} & \multicolumn{1}{c|}{0}& \multicolumn{1}{c|}{0} & \multicolumn{1}{c|}{0} & \multicolumn{1}{c|}{0} & \multicolumn{1}{c|}{0} & \multicolumn{1}{c|}{0} & \multicolumn{1}{c|}{0} & \multicolumn{1}{c}{0}\\ 
 \hline \hline
 \multicolumn{1}{c||}{Overall} &  \multicolumn{1}{c|}{285} & \multicolumn{1}{c|}{129}& \multicolumn{1}{c|}{117} & \multicolumn{1}{c|}{135} & \multicolumn{1}{c|}{120} & \multicolumn{1}{c|}{141} & \multicolumn{1}{c|}{127} & \multicolumn{1}{c|}{142} & \multicolumn{1}{c}{126}\\    \cline{1-10}                    
\end{tabular}
\label{table:isub}
\end{table}
 
\begin{figure*}[!htb]
\begin{center}
\begin{tabular}{cc}
\includegraphics[width=80mm, height=80mm]{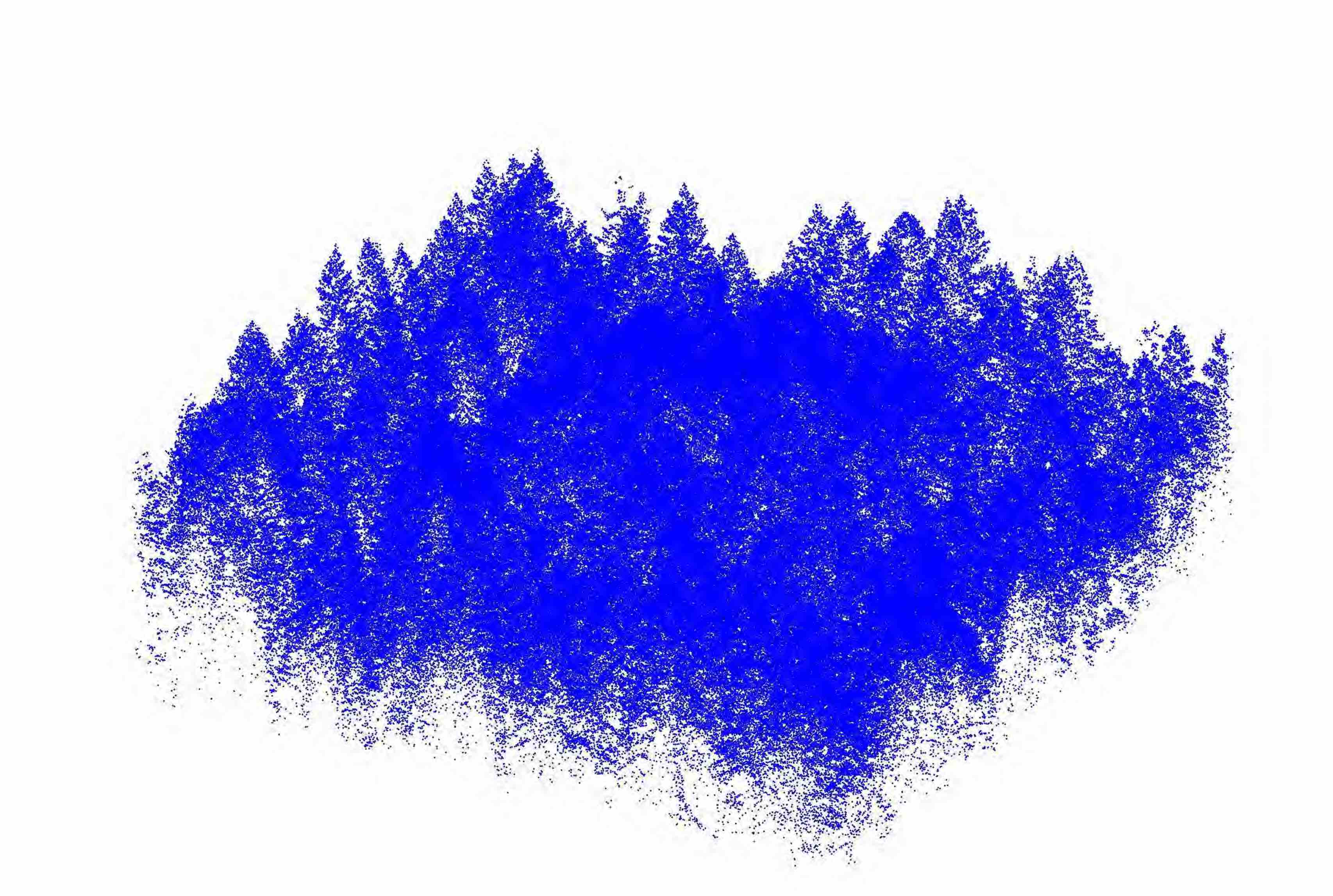}  &
\includegraphics[width=80mm, height=80mm]{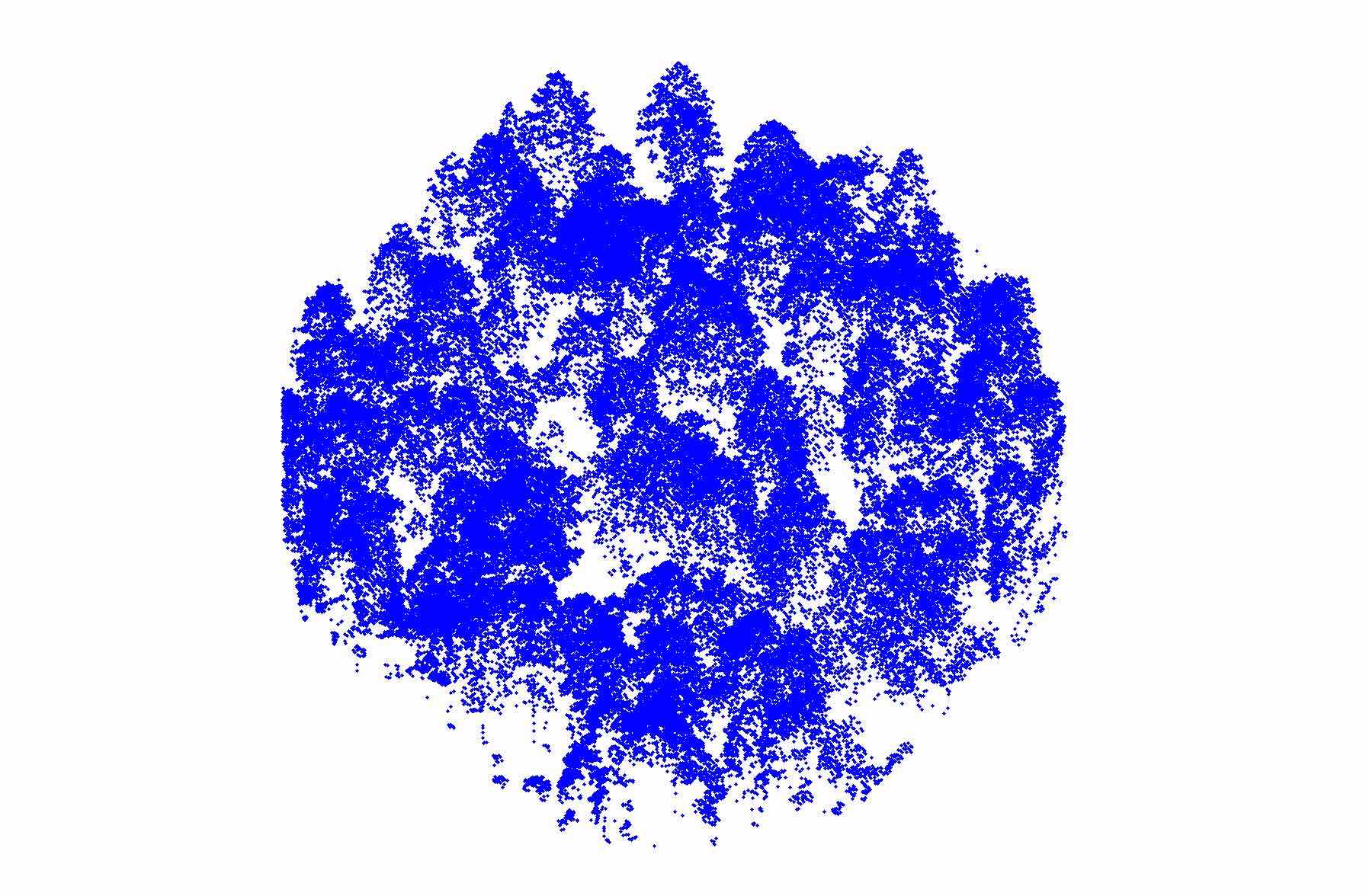}\\
(a) Site 7 & (b) Site 16 \\ 
\includegraphics[width=80mm, height=80mm]{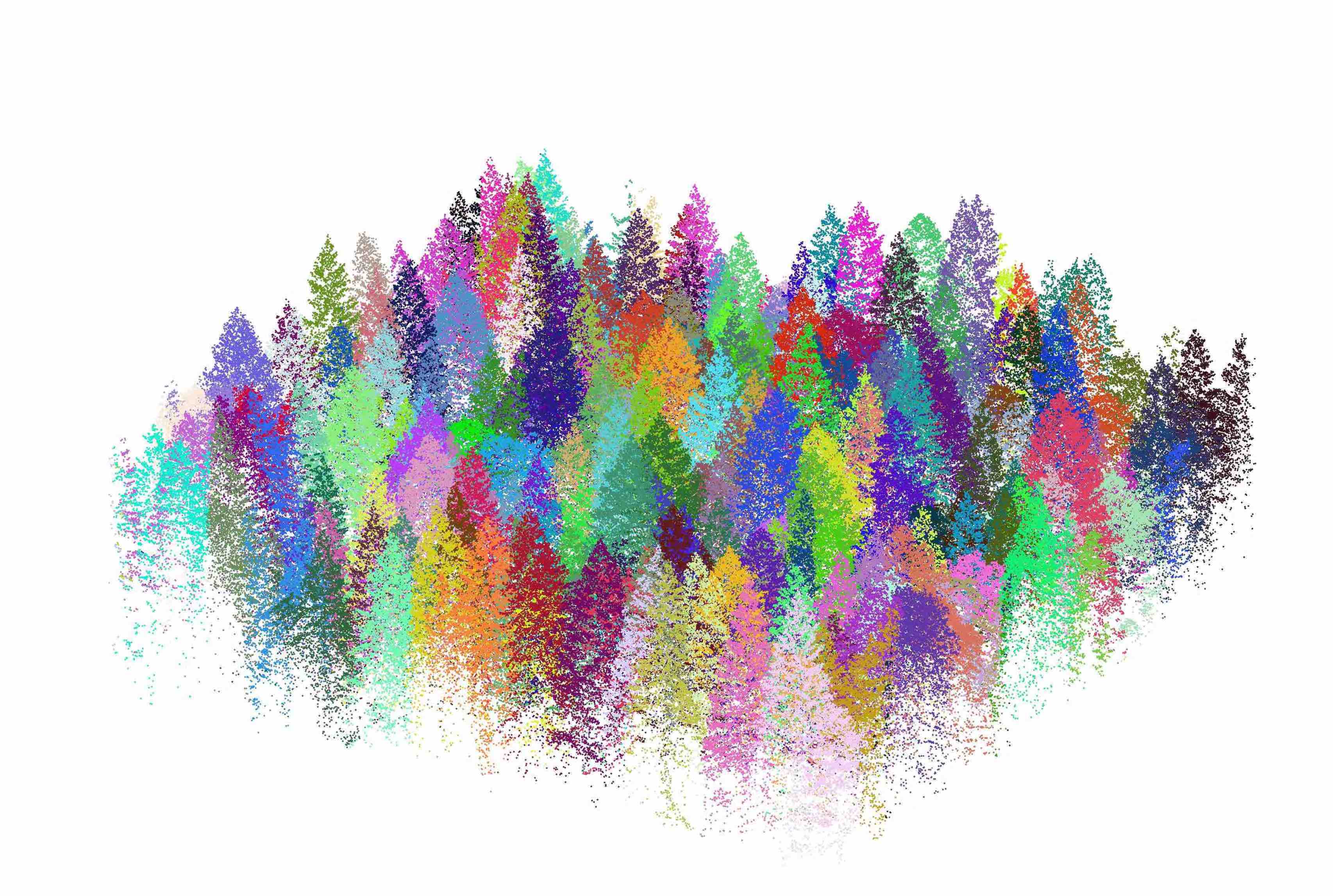}  &
\includegraphics[width=80mm, height=80mm]{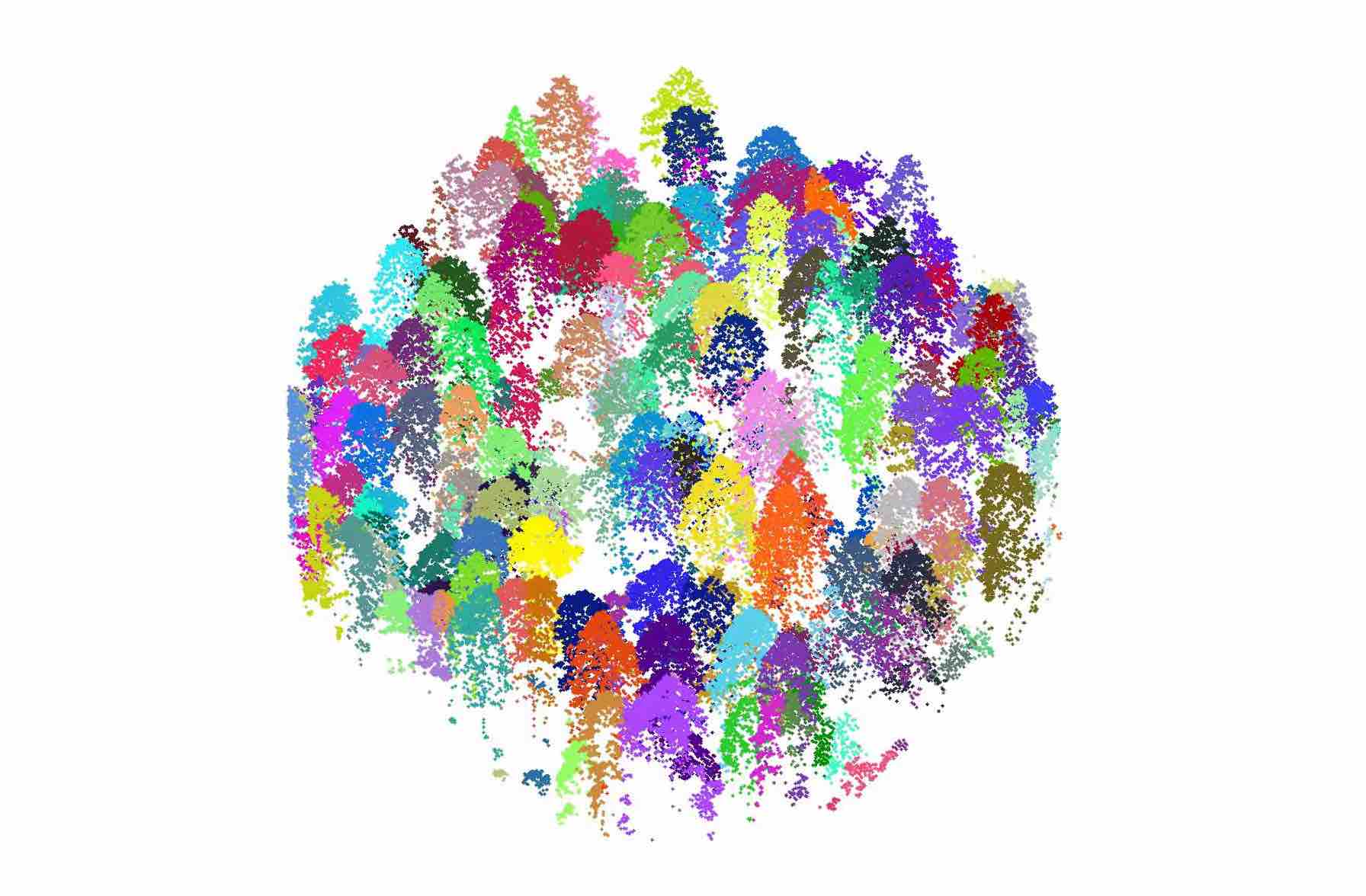} \\
(c) MCRC & (d) MCRC
\end{tabular}
\end{center}
\caption{Examples of MCRC segmentation of the NEWFOR benchmark datasets. The first row shows the LiDAR point clouds from test sites 8 (Left) and 16 (Right). The second row presents the results of the MCRC delineation method,  which assigns each LiDAR return to a tree.}
\label{fig:bench}
\end{figure*}

Further evaluation of the MCRC algorithm (with TIFFS-detected tree top positions as priors) was  conducted using the NewFor benchmark dataset.  The initial tree delineation provided by TIFFS was improved upon by the MCRC segmentation in eight out of fourteen test sites (1, 5, 6, 7, 9, 10, 12 and 16), extracting fewer trees and matching more of them (Table \ref{table:bench}).   In five plots (8, 10, 11, 13 and 18) its performance was similar to that of TIFFs, but performed less well in  one plot (17).  Table \ref{table:bsub} splits these performance figures into different height bands, revealing  that MCRC (a)  reduced the rate of false tree detection and increased the number of trees correctly assigned,  (b) it marginally improved the detection of  small trees;  (c) it over-segmented  trees over 20m in height, as did TIFFS.    Figure \ref{fig:bench} illustrates the segmentation of trees in  study areas 7 and 16 using  MCRC. 

\begin{table}[!h]
\centering 
\caption{Comparison of the performance of TIFFS and MCRC when applied to  NewFor benchmark datasets. `Extract' means the number of trees delineated by the algorithms. `Match' defines the number of correctly assigned trees.} 
\vspace{0.1in}
\footnotesize
\begin{tabular}{cccccc}
\hline
\multicolumn{1}{c||}{Study} & \multicolumn{1}{c|}{Ground}  & \multicolumn{2}{c|}{TIFFS} & \multicolumn{2}{c}{MCRC (TIFFS)} \\ \cline{3-6}
\multicolumn{1}{c||}{ area} &\multicolumn{1}{c|}{truth} & \multicolumn{1}{c|}{Extract} & \multicolumn{1}{c|}{Match} & \multicolumn{1}{c|}{Extract} & \multicolumn{1}{c}{Match}\\ \hline \hline
 \multicolumn{1}{c||}{01} &  \multicolumn{1}{c|}{352} & \multicolumn{1}{c|}{358}& \multicolumn{1}{c|}{181}& \multicolumn{1}{c|}{351}& \multicolumn{1}{c}{188} \\  
 \hline
 \multicolumn{1}{c||}{05} &  \multicolumn{1}{c|}{235} & \multicolumn{1}{c|}{45}& \multicolumn{1}{c|}{41}& \multicolumn{1}{c|}{47}& \multicolumn{1}{c}{44}\\ 
 \hline
 \multicolumn{1}{c||}{06} &  \multicolumn{1}{c|}{47} & \multicolumn{1}{c|}{32}& \multicolumn{1}{c|}{28}& \multicolumn{1}{c|}{34}& \multicolumn{1}{c}{30} \\ 
 \hline
 \multicolumn{1}{c||}{07} &  \multicolumn{1}{c|}{79} & \multicolumn{1}{c|}{61}& \multicolumn{1}{c|}{52}& \multicolumn{1}{c|}{57}& \multicolumn{1}{c}{53} \\ 
 \hline
 \multicolumn{1}{c||}{08} &  \multicolumn{1}{c|}{107} & \multicolumn{1}{c|}{43}& \multicolumn{1}{c|}{39}& \multicolumn{1}{c|}{44}& \multicolumn{1}{c}{38} \\ 
 \hline
 \multicolumn{1}{c||}{09} &  \multicolumn{1}{c|}{169} & \multicolumn{1}{c|}{71}& \multicolumn{1}{c|}{58}& \multicolumn{1}{c|}{67}& \multicolumn{1}{c}{60} \\ 
 \hline
 \multicolumn{1}{c||}{10} &  \multicolumn{1}{c|}{106} & \multicolumn{1}{c|}{79}& \multicolumn{1}{c|}{40}& \multicolumn{1}{c|}{82}& \multicolumn{1}{c}{42} \\                     
 \hline
 \multicolumn{1}{c||}{11} &  \multicolumn{1}{c|}{22} & \multicolumn{1}{c|}{15}& \multicolumn{1}{c|}{10}& \multicolumn{1}{c|}{14}& \multicolumn{1}{c}{10} \\                     
 \hline
 \multicolumn{1}{c||}{12} &  \multicolumn{1}{c|}{49} & \multicolumn{1}{c|}{83}& \multicolumn{1}{c|}{31}& \multicolumn{1}{c|}{76}& \multicolumn{1}{c}{32} \\                     
 \hline
 \multicolumn{1}{c||}{13} &  \multicolumn{1}{c|}{100} & \multicolumn{1}{c|}{63}& \multicolumn{1}{c|}{45}& \multicolumn{1}{c|}{62}& \multicolumn{1}{c}{45} \\                     
 \hline
 \multicolumn{1}{c||}{15} &  \multicolumn{1}{c|}{53} & \multicolumn{1}{c|}{42}& \multicolumn{1}{c|}{24}& \multicolumn{1}{c|}{41}& \multicolumn{1}{c}{23} \\                     
 \hline
 \multicolumn{1}{c||}{16} &  \multicolumn{1}{c|}{37} & \multicolumn{1}{c|}{45}& \multicolumn{1}{c|}{21}& \multicolumn{1}{c|}{42}& \multicolumn{1}{c}{23} \\                     
 \hline
 \multicolumn{1}{c||}{17} &  \multicolumn{1}{c|}{117} & \multicolumn{1}{c|}{82}& \multicolumn{1}{c|}{69}& \multicolumn{1}{c|}{80}& \multicolumn{1}{c}{64} \\                     
 \hline
 \multicolumn{1}{c||}{18} &  \multicolumn{1}{c|}{92} & \multicolumn{1}{c|}{58}& \multicolumn{1}{c|}{42}& \multicolumn{1}{c|}{62}& \multicolumn{1}{c}{42} \\  \hline \hline
 \multicolumn{1}{c||}{Overall} &  \multicolumn{1}{c|}{1565} & \multicolumn{1}{c|}{1074}& \multicolumn{1}{c|}{681}& \multicolumn{1}{c|}{1060}& \multicolumn{1}{c}{695} \\   \cline{1-6}                  

\end{tabular}
\label{table:bench}
\end{table}

\begin{table}[!h]
\centering 
\caption{The summary of the performance of delineation algorithms in the NewFor benchmark dataset in different tree height tiers. `Extract' means the number of trees delineated by the algorithms. `Match' is the number of trees that were matched to trees in the mapped forest plot which had similar (x,y) coordinates and were of similar heights}
\vspace{0.1in}
\footnotesize
\begin{tabular}{cccccc}
\hline
\multicolumn{1}{c||}{Study} & \multicolumn{1}{c|}{Ground}  & \multicolumn{2}{c|}{TIFFS} & \multicolumn{2}{c}{MCRC (TIFFS)} \\ \cline{3-6}
\multicolumn{1}{c||}{ area} &\multicolumn{1}{c|}{truth} & \multicolumn{1}{c|}{Extract} & \multicolumn{1}{c|}{Match} & \multicolumn{1}{c|}{Extract} & \multicolumn{1}{c}{Match}\\ \hline \hline
 \multicolumn{1}{c||}{$ h \geq 20\text{m}$} &  \multicolumn{1}{c|}{638} & \multicolumn{1}{c|}{811}& \multicolumn{1}{c|}{547}& \multicolumn{1}{c|}{797}& \multicolumn{1}{c}{550} \\  
 \hline
 \multicolumn{1}{c||}{$ 15\text{m}\leq h < 20\text{m}$} &  \multicolumn{1}{c|}{279} & \multicolumn{1}{c|}{147}& \multicolumn{1}{c|}{96}& \multicolumn{1}{c|}{155}& \multicolumn{1}{c}{97}\\ 
 \hline
 \multicolumn{1}{c||}{$ 10\text{m}\leq h < 15\text{m}$} &  \multicolumn{1}{c|}{292} & \multicolumn{1}{c|}{34}& \multicolumn{1}{c|}{21}& \multicolumn{1}{c|}{40}& \multicolumn{1}{c}{27} \\ 
 \hline
 \multicolumn{1}{c||}{$ 5\text{m}\leq h < 10\text{m}$} &  \multicolumn{1}{c|}{270} & \multicolumn{1}{c|}{41}& \multicolumn{1}{c|}{14}& \multicolumn{1}{c|}{41}& \multicolumn{1}{c}{18} \\ 
 \hline
 \multicolumn{1}{c||}{$ 2\text{m}\leq h < 5\text{m}$} &  \multicolumn{1}{c|}{86} & \multicolumn{1}{c|}{41}& \multicolumn{1}{c|}{3}& \multicolumn{1}{c|}{27}& \multicolumn{1}{c}{3} \\ 
            
  \hline \hline
 \multicolumn{1}{c||}{Overall} &  \multicolumn{1}{c|}{1565} & \multicolumn{1}{c|}{1074}& \multicolumn{1}{c|}{681}& \multicolumn{1}{c|}{1060}& \multicolumn{1}{c}{695} \\   \cline{1-6}                  

\end{tabular}
\label{table:bsub}
\end{table}

Neither TIFFS and MCRC were very successful at delineating trees within the English broadleaf dataset, but  MCRC outperformed TIFFS. Broadleaf trees have less distinctive tree tops than the conifers of  Italian and Alpine datasets, making delineation more of a challenge.  Overall, MCRC extracted 346 trees and correctly matched 197 trees, showing that 14 more trees were correctly segmented than that by TIFFS.    However,  this amounts to only 8--10 percent of trees (first row of Table \ref{table:wytham} ),  and virtually none of the  trees  under 15m in height  were found (bottom three rows of Table \ref{table:wytham} ). This is in part due to the low point density of the LiDAR dataset, which makes it particularly  challenging to find small trees.  The analysis of trees > 20 m tall shows that both TIFFS and MCRC over-segmented ITCs. However, the ratio between extracted and matched canopy trees was 50.3\% and 54.8\%, respectively, indicating that false positives were reduced by 4.5\% when using MCRC. 
 
Table \ref{table:exp-time} compares the computational time for the RC and MCRC with TIFFS,  when applied to the Italian datasets (plots of 15m radius with unusually high point density). As RC needs to construct a graph recursively to segment trees, its computational cost is more expensive than that of TIFFS or MCRC on this high point-density dataset. RC was ten times slower than MC and twice slower than MCRC, because it  separates point clouds into only two clusters at each step.  

\begin{table}[!h]
\centering 
\caption{The performance of the delineation algorithms in the English dataset, by height tier. `Extract' means the number of delineated trees. `Match' is the number of trees that were matched to trees in the mapped forest plot which had similar (x,y) coordinates and similar heights. In the first column,  the range of heights in each tier is shown.}
\vspace{0.1in}
\footnotesize
\begin{tabular}{cccccccc}
\hline
\multicolumn{1}{c||}{Study area:} & \multicolumn{1}{c|}{Ground}  & \multicolumn{2}{c|}{TIFFS} & \multicolumn{2}{c|}{MCRC (TIFFS)} & \multicolumn{2}{c}{MCRC (TIFFS) Hyp} \\ \cline{3-8}
\multicolumn{1}{c||}{Wytham} &\multicolumn{1}{c|}{truth} & \multicolumn{1}{c|}{Extract} & \multicolumn{1}{c|}{Match} & \multicolumn{1}{c|}{Extract} & \multicolumn{1}{c|}{Match} & \multicolumn{1}{c|}{Extract} & \multicolumn{1}{c}{Match}\\ \hline \hline
 \multicolumn{1}{c||}{$h>$ 0m} &  \multicolumn{1}{c|}{2116} & \multicolumn{1}{c|}{342}& \multicolumn{1}{c|}{183}& \multicolumn{1}{c|}{346}& \multicolumn{1}{c|}{197} & \multicolumn{1}{c|}{419}& \multicolumn{1}{c}{225} \\ \hline \hline
 \multicolumn{1}{c||}{$h \geq $20m} &  \multicolumn{1}{c|}{194} & \multicolumn{1}{c|}{280}& \multicolumn{1}{c|}{141 }& \multicolumn{1}{c|}{264}& \multicolumn{1}{c|}{147 } & \multicolumn{1}{c|}{318}& \multicolumn{1}{c}{166 } \\  \hline
  \multicolumn{1}{c||}{$15\text{m}\leq h < 20\text{m}$} &  \multicolumn{1}{c|}{476} & \multicolumn{1}{c|}{61}& \multicolumn{1}{c|}{41 }& \multicolumn{1}{c|}{76}& \multicolumn{1}{c|}{50} & \multicolumn{1}{c|}{96}& \multicolumn{1}{c}{58} \\  \hline
   \multicolumn{1}{c||}{ $10\text{m}\leq h < 15\text{m}$} &  \multicolumn{1}{c|}{523} & \multicolumn{1}{c|}{1}& \multicolumn{1}{c|}{1}& \multicolumn{1}{c|}{4}& \multicolumn{1}{c|}{0} & \multicolumn{1}{c|}{3}& \multicolumn{1}{c}{1} \\  \hline
    \multicolumn{1}{c||}{$5\text{m}\leq h < 10\text{m}$} &  \multicolumn{1}{c|}{756} & \multicolumn{1}{c|}{0}& \multicolumn{1}{c|}{0}& \multicolumn{1}{c|}{2}& \multicolumn{1}{c|}{0} & \multicolumn{1}{c|}{2}& \multicolumn{1}{c}{0} \\  \hline
        \multicolumn{1}{c||}{$2\text{m}\leq h < 5\text{m}$} &  \multicolumn{1}{c|}{159} & \multicolumn{1}{c|}{0}& \multicolumn{1}{c|}{0}& \multicolumn{1}{c|}{0}& \multicolumn{1}{c|}{0} & \multicolumn{1}{c|}{0}& \multicolumn{1}{c}{0}   \\  \hline
\end{tabular}
\label{table:wytham}
\end{table}

\subsection{Tree delineation from LiDAR and hyperspectral imagery}

MCRC provides a framework for using both LiDAR point cloud and features from hyperspectral imagery, and we tested this approach with the Italian and English datasets (far right columns in Tables 2,3 and 6). For the Italian dataset (Tables \ref{table:performance} and \ref{table:isub})  hyperspectral imagery does not improve the already excellent segmentation of upper canopy trees. In plot 77, 16 out of 18 trees were correctly matched compared to 17 out of 19 trees in the LiDAR-only analysis. In plot 102, fewer false positive were detected than the LiDAR-only analysis, while more false positive were detected in plots 129 and 292. No difference was noticed in plots 91, 220 and 274. 
When  LiDAR and hyperspectral imagery were used in the MCRC to detect trees in  English broadleaf forest,  more trees were detected than with TIFFS or the LiDAR-only graph cut. Examining only the  canopy trees over 20m height (9\% of recorded trees), the number of extracted and correctly assigned trees  increased by 38 and 19, respectively. However, these  large trees were over-segmented :   194  were observed in the ground data, but  the full MCRC identified  318 trees, of which 166 trees were  matched. The ratio of extracted to matched tall trees for TIFFS, MCRC with LiDAR only and MCRC with LiDAR and hyperspectral imagery was 50.3\%, 54.8\% and 52.2\%, respectively, indicating that more trees were detected but at the expense of more false positives when all the remote sensing information was fused. The detection of the understory trees ($<$ 15m) in Table \ref{table:wytham} was poor for all algorithms. The full MCRC found eight more trees than MCRC with LiDAR only and 17 more trees than TIFFS in the range of tree heights in $10\text{m}< x < 15\text{m}$ , but at the expense of a larger commission error.

\begin{table}[!h]
\centering \caption{Computation time in seconds of RNC, MCRC, TIFFS and MCRC applied to the Italian dataset.}
\vspace{0.1in}
\begin{tabular}{c||c|c|c|c}
\hline
  \multicolumn{1}{r||}{}  &  \multicolumn{1}{c|}{RNC}  & \multicolumn{1}{c|}{MCRC} & 
  \multicolumn{1}{c|}{MCRC} & \multicolumn{1}{c}{TIFFS} \\ \hline \hline
  \multirow{1}{*}{Plot 77}  &  486 &  39  & 227 & 12\\  \hline
  \multirow{1}{*}{Plot 91}  &  621  & 57 & 227  & 12 \\  \hline
  \multirow{1}{*}{plot 102} &  888 & 79 & 494 & 19
 \\  \hline
  \multirow{1}{*}{plot 129} &  417 & 48 &  276 & 12\\  \hline
  \multirow{1}{*}{plot 220} &  1085 & 103 & 809 & 14\\  \hline
  \multirow{1}{*}{plot 274} &  686 & 84 &  320 & 12\\  \hline
  \multirow{1}{*}{plot 292} &  653 & 52 & 377 &12 \\ \hline
\end{tabular}
\label{table:exp-time}
\end{table}

\subsection{Extraction of tree properties  from delineated point clouds}

\begin{figure*}[!htb]
\begin{center}
\begin{tabular}{cc}
\includegraphics[width=75mm, height=75mm]{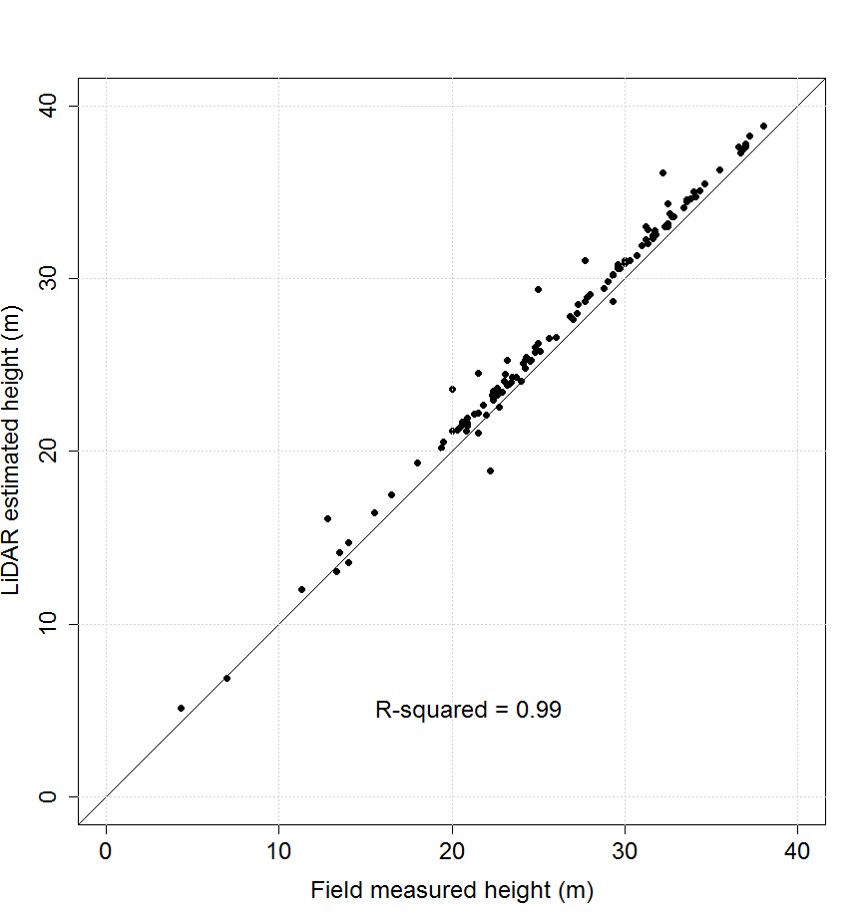}  &
\includegraphics[width=75mm, height=75mm]{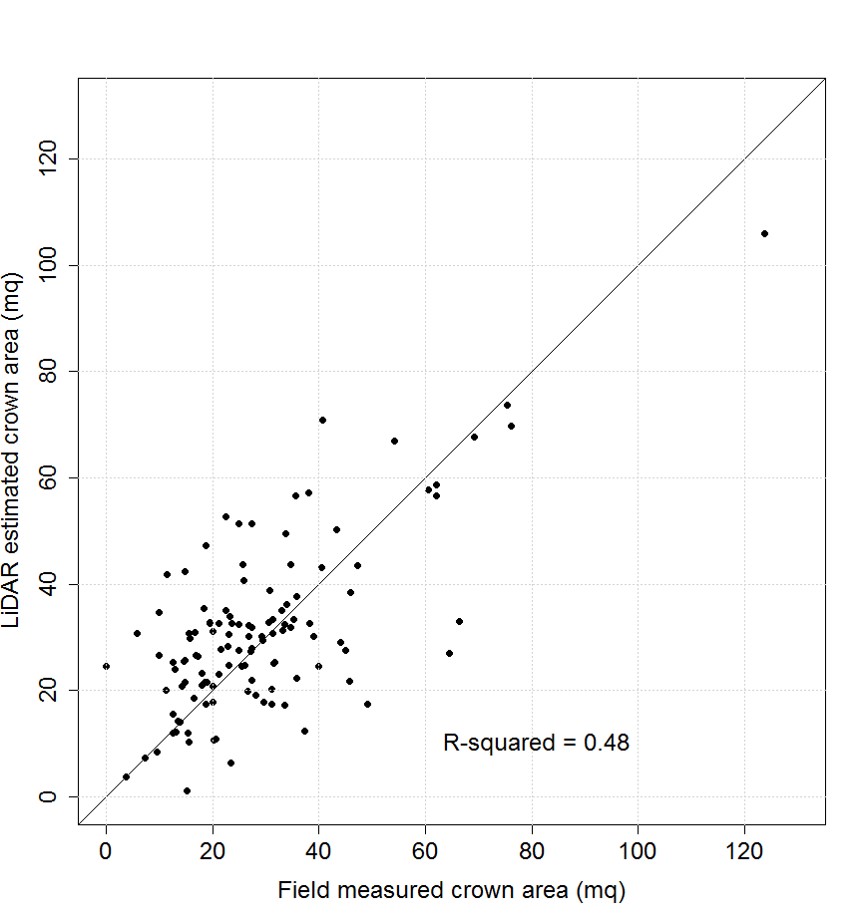}\\
(a) LiDAR -- Field measured tree height & (b) LiDAR -- Field measured crown area \\ 
\end{tabular}
\end{center}
\caption{Scatterplots of estimated tree heights (a) and crown areas (b) extracted  using the MCRC algorithm, compared with values estimated on the ground in the Italian Alps}
\label{fig:est}
\end{figure*}

MCRC was successfully used to extract information from individual trees .  We selected trees from the Italian Alps dataset for which  a match had been found between delineated trees and ground information. The tree height estimation was nearly perfect \ref{fig:est} , but this result needs to be treated with caution, as the  tree matching algorithm (in NewFor validation software) uses tree heights as a variable to match segmented trees with those of the ground.  The relationship between field and remotely sensed crown area (\ref{fig:est}) is stronger than obtained by us previously ( Dalponte and Coomes, \textit{in press}) - the R-squared value of the regression relationship was 0.48.

\section{Discussion}\label{sec:discussion}

\subsection{The application of graph cut approaches to tree delineation}
The multi-class normalised cut approach, constrained with information from classical CHM-based delineation, improved the quality of ITC  delineation, although not to the degree we had hoped. The validity of MCRC was demonstrated by experiments using the NewFor benchmark, English and Italian datasets, which shows that it outperformed a leading CHM-based segmentation algorithm in most cases.   It also outperformed the  RC method, perhaps because RC discards too much useful information by working with only the second smallest eigenvector \citep{shi2000normalized,von2007tutorial}. 
	We used strong priors which strongly influenced the tree detection accuracy of the MCRC.  The algorithm can successfully detect more trees than predicted by the number of local maxima  provided as a prior,  because the RC step provides opportunity for  further separation of each cluster.  But, there is hardly any  opportunity to merge clusters identified in the prior.  For example, when TIFFS incorrectly detected four tree tops in plot 129 of the Italian dataset, this leads to those same trees being detected by MCRC.  Merging of over-segmented trees can occur if the local maxima are so close together that the point clouds coalesce when generating the prior. This is indeed seen with the English broadleaf dataset,  where MCRC  extracted 14 fewer trees than  TIFFS,  because several of the local maxima identified by TIFFS were close together and merged into a single cluster.   But a "softer" constraint would improve the performance of the MCRC.  With a  "soft'' constraint, the correlation need not be satisfied, but instead the algorithm finds a balance between the maximal correlation and optimal normalised cut separation.  This would require us to build a new optimisation model, which is beyond the scope of this study. 
 
\subsection{Combining LiDAR and hyperspectral imagery to improve delineation}

MCRC was able to detect more understory trees than CHM-based approaches,  but could not find the small trees under dense forests.  In principle, it should be able to detect understory trees if the point density of LiDAR data was high enough to represent understory structures. In case of the English dataset, the point density was only 6  m$^-2$, with few points penetrating the upper canopy, making it hard to find any understory structure. In contrast, the LiDAR point density of the Italian dataset was so high that internal structures of trees, and understory trees could be identified, which may explain why MCRC performed better than TIFFS in this case. However, even with this dataset there were still a number of understory trees undetected by MCRC.   Considering that we used a fixed set of parameters for all benchmark testing, the performance of MCRC could probably be improved with manual parameter tuning.

If we consider vertical LiDAR point profiles of each canopy, we can change the parameters for the RC step. Duncanson {\it et al.} \citep{duncanson2014efficient} used the vertical distributions of LiDAR point clouds to separate understory trees from taller individuals. After an initial ITC delineation using a watershed algorithm the authors examined the vertical LiDAR point distribution to see whether it showed continuous decrease from the top canopy or whether there were through in the distribution, indicating separation between understory and canopy trees. This approach can be applied to our segmentation algorithm directly or the vertical profiles can be parameterised to be incorporated into the RC step. Also full-waveform LiDAR may provide an opportunity to find internal structures in more details. Reitberger {\it et al.} used RC with  full-waveform LiDAR, which had 9 points per m$^2$ \citep{reitberger20093d} . They  suggested full-waveform LiDAR pulse and intensity with calibration could help to detect ITCs in the understory. Unfortunately, LiDAR intensity was not calibrated in our datasets due to an automatic gain control system on the Leica instrument, which regulates LiDAR intensities in non-linear and opaque way,  so intensity could not be used in the segmentation.

\begin{figure*}[!htb]
\begin{center}
\begin{tabular}{cc}
\includegraphics[width=75mm, height=75mm]{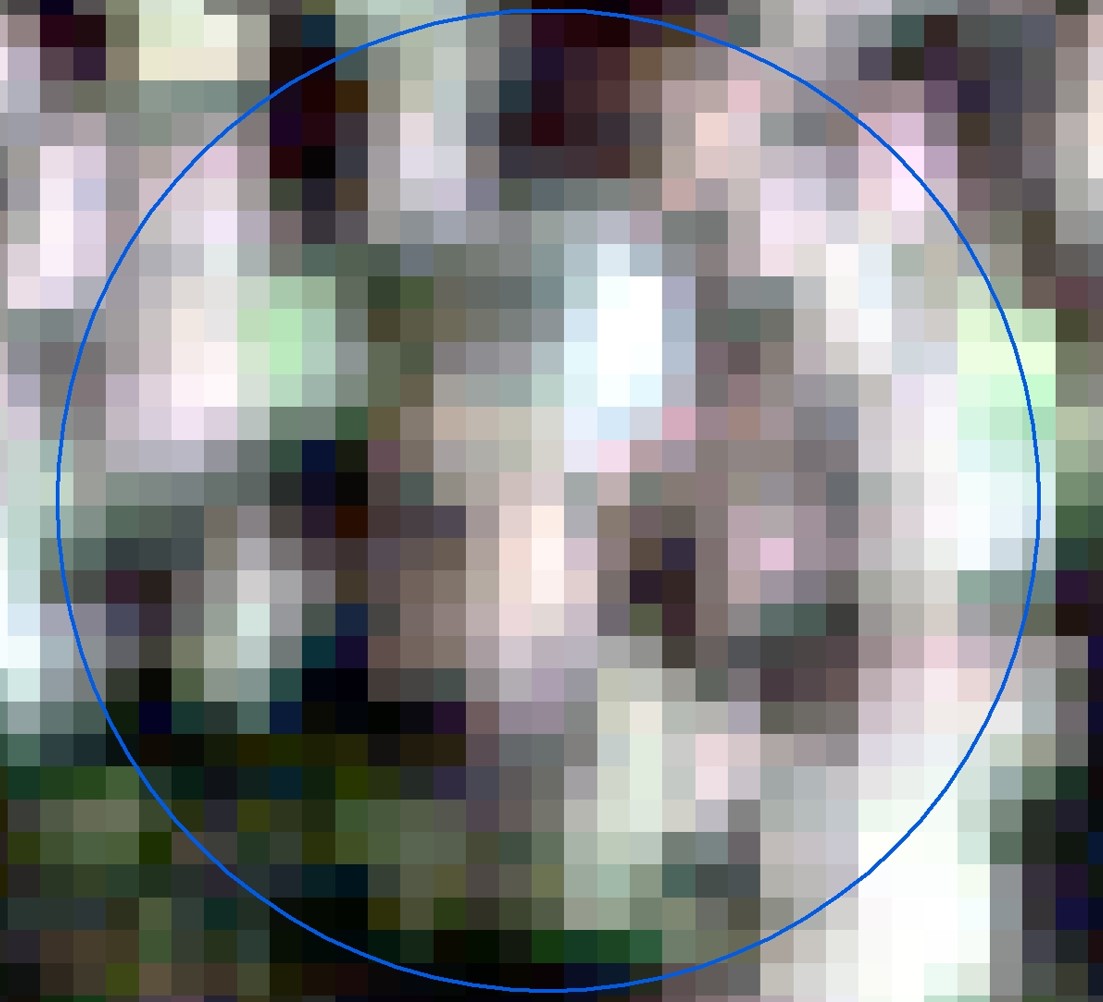}  &
\includegraphics[width=75mm, height=75mm]{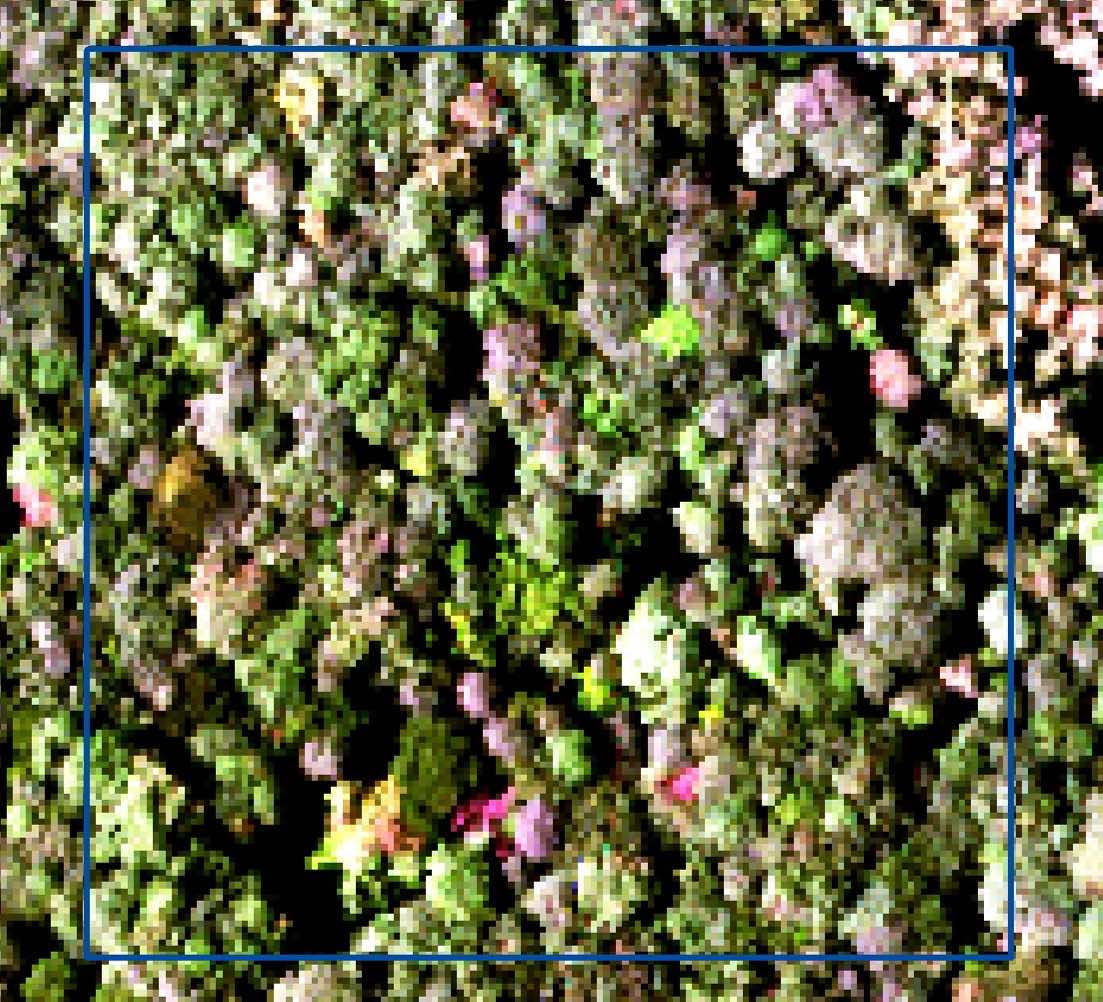}\\
(a) Hyperspectral imagery of the Italian dataset (plot 220) & (b) Hyperspectral imagery of the English dataset \\ 
\end{tabular}
\end{center}
\caption{Examples of hyperspectral images in the Italian and English datasets. The blue circle and square represent the size of test sites of the Italian and English datasets, respectively. The spatial resolution of hyperspectral images are 1m and 1.2m, respectively.}
\label{fig:hyp}
\end{figure*}

The experiments using both LiDAR and hyperspectral data in the Italian Alps showed that the ITC delineation was not improved  by hyperspectral imagery, while those of the English dataset improved the number of trees correctly segmented at the expense of greater over-segmentation.  Figure \ref{fig:hyp} shows the hyperspectral images of the Italian and English datasets.  As shown in Figure \ref{fig:hyp}(a), the pixel size of the hyperspectral imagery in the Italian dataset was too large to give precise feature information to segment dense LiDAR point clouds ($\geq$ 80 points per $\text{m}^2$). LiDAR point density was very high - almost hundred points were represented by a single hyperspectral pixel. Under this condition, ITC delineation is mainly driven by LiDAR point cloud rather than hyperspectral imagery. As  the Italian plots were often  dominated by just two species, the information provided by the hyperspectral imagery was not useful for the ITC delineation. In the English dataset, on the other hand, LiDAR point density was low (6 points per m$^2$) and there was a higher species diversity (see Figure \ref{fig:hyp}(b)). These two conditions made the English dataset much better for ITC delineation using both types of imagery. However, more false positive were observed when both LiDAR and hyperspectral imagery were used in the MCRC. This may be related to shade effects or registration errors that remained in the hyperspectral imagery.  It was reported that the illumination effects contained in the first principal component of the hyperspectral imagery cause inaccurate ITC delineation \citep{tochon2015use},  so we extracted 2--5th principal components of hyperspectral imagery for ITC delineation.  However, the illumination effects may still remain in the principal components we used for the delineation \citep{tochon2015use}.

\subsection{The problem of detecting understory trees}

\begin{figure*}[!htb]
\begin{center}
\begin{tabular}{c}
\includegraphics[width=80mm, height=80mm]{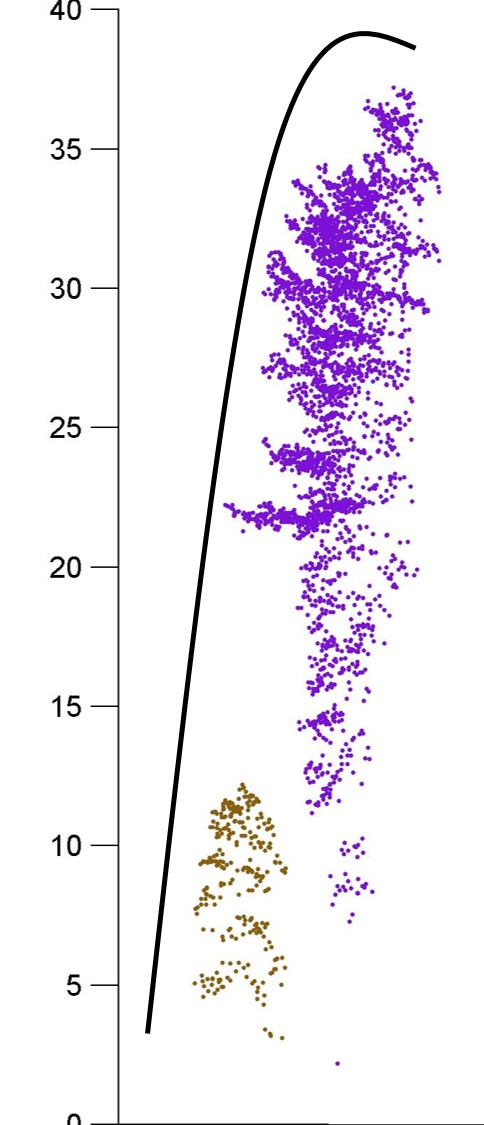}\\
\end{tabular}
\end{center}
\caption{Example of MCRC segmentation of understory tree. The black solid line is the interpolation line (CHM) of the LiDAR point cloud. Point clouds in sienna and purple colours are the segmented ITCs using MCRC.}
\label{fig:understory}
\end{figure*}

The detection of understory trees is strongly influenced by the point density. In the English dataset, the relatively low point density (6 points per m$^2$) made it was impossible to detect understory trees, thus causing very low detection rates.  Low detection rate may also be attributable to uncertainties in the locations of trees on the ground, which meant that matches were not made by the NewFor algorithm even though delineation had been accurate. By contrast,  the point density was extremely high in the Italian dataset, making it was possible to extract understory tree because there is more information regarding small trees in the cloud. Figure \ref{fig:understory} shows an example of understory tree delineation using MCRC. In this example, in the CHM (black solid line) only a single tree crown was visible, as the CHM is constructed by the interpolation of LiDAR point cloud. In contrasts, MCRC can delineate two ITCs, because the RC process checks for further separability of each ITCs and the LiDAR point density was high enough to find understory it. In this example, the understory tree was clearly separable because there was a vertical gap between trees. However, this is not common in dense LiDAR point cloud because canopy and understory trees usually overlap. In this case, parameters for graph weights should be chosen carefully. Since we fixed the parameters for the RC process for all the ITCs, it is hard to delineate subcanopy trees efficiently. LiDAR vertical profiles of each canopy tree may provide good statistics for separating understory trees \citep{duncanson2014efficient}. If we can learn parameters automatically from LiDAR vertical statistics, then ITC delineation can be extended to find understory trees. However, analysing LiDAR vertical profiles and learning ITC parameters are beyond the scope of our research. 

\subsection{Concluding remarks}

This paper has described a  normalised cut approach for  ITC delineation. Our experiments show that MCRC outperforms convention CHM-based techniques  in all test datasets.   MCRC easily incorporates optical imagery alongside LiDAR data, so ITC delineation could be conducted using LiDAR and optical imagery.  Since MCRC assigns each LiDAR object return to a tree, it can be used to measure tree dimensions accurately.  Constructing a large graph and solving eigensystem repeatedly is costly in terms of computational time,  but MCRC separates LiDAR point cloud into clusters during the initial MC step, so the graph size of each segment is relatively small for the recursive binary step. In addition, parallelisation can be implemented for RC step because we can apply the algorithm to each segment.
	The truth is, though, that the  slightly superior performance of MCRC over  classic CHM-based approaches, does not warrant its widespread use at this time, because the computation costs are high and the benefits small.  We see a number of ways  in which it could be improved though.  Watershed algorithm tend to over-segment large trees,  and this over-segmentation works cannot be reversed by our graph cut algorithm which uses these tree locations as priors.  Replacing our hard constraint with a softer one may resolve this problem, and would the use of delineation approaches that are less prone to over-segmentation.   MCRC is  computationally expensive -  if we have thousands of local maxima then we need to compute  thousands of eigenvectors to delineate the ITCs, which increases the memory complexity. This problem can be avoided by domain decomposition and parallelisation techniques.   Understory trees should  be detected by RC process if the LiDAR point density is high enough, but more work is needed to refine the approach.  Combining our method with the multilayer detection approach of  \citep{duncanson2014efficient} could be particularly fruitful. Despite these limitations, by making full use of the data available, graph cut has the potential to considerably improve the accuracy of tree delineation.

\noindent{\bf Acknowledgments.} The authors would like to thank NERC-ARSF and the data analysis node for collecting and pre-processing the Wytham Woods dataset used in this research project [RG13/08/175b]. Xiaohao Cai was supported by the Issac Newton Trust and Welcome Trust. DAC was supported by a DEFRA-BBSRC grant to study the spread of ash dieback disease in British woodlands. 

\bibliographystyle{apalike}

\bibliography{segment}

\end{document}